\documentclass[10pt,twocolumn,letterpaper]{article}

\usepackage{iccv}
\usepackage{times}
\usepackage{epsfig}
\usepackage{graphicx}
\usepackage{amsmath}
\usepackage{amssymb}
\usepackage{array}

\usepackage{subfigure}     
\usepackage{multirow}       
\usepackage{booktabs}       
\usepackage{float}
\usepackage[graphicx]{realboxes}
\usepackage{tabularx}
\usepackage{pifont}
\usepackage{algorithm}
\usepackage{xspace}

\usepackage{color, colortbl}
\definecolor{Gray}{gray}{0.9}
\definecolor{Goldenrod}{RGB}{230, 240, 255}%
\usepackage{xcolor}

\newcommand{\cmark}{\ding{51}\xspace}%
\newcommand{\xmark}{\ding{55}\xspace}%
\newcommand{\xmarkg}{\textcolor{lightgray}{\ding{55}}\xspace}%

\usepackage{mathtools}
\usepackage{amsthm}

\usepackage[breaklinks=true,bookmarks=false]{hyperref}

\iccvfinalcopy 

\begin{document}

\title{Rethinking Evaluation Protocols of Visual Representations Learned via Self-supervised Learning}

\author{%
  Jae-Hun Lee$^{1\thanks{Equal contribution}}$\ \ \
  Doyoung Yoon$^{1\footnotemark[1]}$\ \ \ 
  ByeongMoon Ji$^{1\footnotemark[1]}$\ \ \
  Kyungyul Kim$^{1}$\ \ \
  Sangheum Hwang$^{1,2}$\thanks{Corresponding author}\\
  \\
  \small  $^{1}$LG CNS AI Research, Seoul, South Korea\\
  \small  $^{2}$Department of Data Science, Seoul National University of Science and Technology, Seoul, South Korea\\
  \tt\small\texttt{\{Jaehun.Lee, dy0916, jibm, kyungyul.kim, shwang\}@lgcns.com}
}

\maketitle
\ificcvfinal\thispagestyle{empty}\fi

\begin{abstract}
Linear probing (LP) (and $k$-NN) on the upstream dataset with labels (e.g., ImageNet) and transfer learning (TL) to various downstream datasets are commonly employed to evaluate the quality of visual representations learned via self-supervised learning (SSL). Although existing SSL methods have shown good performances under those evaluation protocols, we observe that the performances are very sensitive to the hyperparameters involved in LP and TL. We argue that this is an undesirable behavior since truly generic representations should be easily adapted to any other visual recognition task, i.e., the learned representations should be robust to the settings of LP and TL hyperparameters. In this work, we try to figure out the cause of performance sensitivity by conducting extensive experiments with state-of-the-art SSL methods. First, we find that input normalization for LP is crucial to eliminate performance variations according to the hyperparameters. Specifically, batch normalization before feeding inputs to a linear classifier considerably improves the stability of evaluation, and also resolves inconsistency of $k$-NN and LP metrics. Second, for TL, we demonstrate that a weight decay parameter in SSL significantly affects the transferability of learned representations, which cannot be identified by LP or $k$-NN evaluations on the upstream dataset. We believe that the findings of this study will be beneficial for the community by drawing attention to the shortcomings in the current SSL evaluation schemes and underscoring the need to reconsider them.
\end{abstract}

\section{Introduction}
\label{sec:intro}
Self-supervised learning (SSL) has emerged as a promising approach for learning generic visual representations with large amounts of unlabeled data. To learn such useful representations with unlabeled data, pretext tasks are typically defined as proxies for training objectives. Many studies have proposed effective learning strategies: contrastive learning that performs instance discrimination based on randomly augmented views~\cite{UFL, SimCLR, MoCo_v3, AMDIM}, a teacher-student framework that trains representations by using outputs of a momentum encoder as supervision~\cite{BYOL, OBoW, DINO, EsViT}, and masked image modeling~\cite{MAE, BEiT, MST, iBOT} that aims to reconstruct randomly masked patches. 

In previous studies, the quality of representations learned via SSL has been evaluated in terms of \textit{discriminability} and \textit{transferability}. The former focuses on how well high-level semantics of pretraining data (i.e., upstream data) are discriminated on the learned representation space. The latter addresses whether the learned representations are generally useful for other visual recognition datasets (i.e., downstream data). Ideally, it is desirable that truly generic representations have both of these properties: they should capture object-level concepts of upstream data, while being transferable to other downstream tasks.

Commonly used schemes for the evaluation of \textit{discriminability} are linear probing (LP, a.k.a. linear evaluation) and $k$-NN~\cite{DINO}. They perform linear or $k$-NN classification with the learned representations frozen. Although these schemes are intuitive and widely accepted in the community, we observe that in the case of LP, considerable effort (e.g., hyperparameter tuning) has been devoted to training such a simple linear classifier. In other words, LP performance is very sensitive to the hyperparameters (results shown in Section~\ref{sec:method1}). Furthermore, the rank of the reported performance measured by LP and $k$-NN does not match (refer to Figure 1 in~\cite{Beyond}). This makes it difficult to compare the discriminability of representations learned by existing SSL methods.

To evaluate \textit{transferability}, transfer learning (TL), i.e., finetuning (FT) a pretrained SSL backbone with various downstream datasets, is typically employed. One can expect good classification performance across datasets if the pretrained features are considered as general-purpose visual representations. However, similar to LP, it is observed that the TL performance of the SSL backbone varies significantly depending on the hyperparameters involved in FT while that of a supervised pretrained backbone does not (results shown in Section~\ref{sec:method2}). Since existing SSL methods are highly tuned by exploring a substantially large hyperparameter space, it is challenging to determine which SSL method can provide useful representations with broad applicability across diverse downstream tasks.

We claim that truly generic representations should be easily transferable and applicable to other visual recognition tasks. From a technical point of view, pretrained representations must be insensitive to the setting of hyperparameters employed in the evaluation protocols (i.e., LP and TL). 
To examine hyperparameter sensitivity, we conduct extensive experiments under cross-settings of hyperparameters. For example, we use LP or TL setting of DINO~\cite{DINO} to evaluate the corresponding performance of the MoCo v3~\cite{MoCo_v3} model, and vice versa. Experimental results reveal that both LP and TL show a large performance gap depending on the setting of hyperparameters. Then, \textit{which factors contribute to the sensitivity of performance? Does this imply that the quality of representations from current SSL methods does not yet meet our expectations? Or are we missing something in the current evaluation protocols?} In this work, we address these questions. It should be noted that our goal is not to compare the performance of existing SSL methods, but rather to identify the source of performance sensitivity under current evaluation schemes.

For LP, we find that distributions of representations obtained from each SSL method show a significant difference. A missing component in most previous SSL studies (except MAE~\cite{MAE}) is input normalization although it is a basic and indispensable preprocessing step for effective training. Our observations indicate that applying batch normalization (BN) before feeding inputs (i.e., SSL representations) to a linear classifier results in nearly eliminating performance variations across different hyperparameter settings. Furthermore, the inconsistency issue between LP and $k$-NN evaluations is resolved: a method demonstrating superior LP performance (with BN) also yields better results in $k$-NN evaluation.
In TL through FT, it is identified that there exists a hidden hyperparameter in the upstream stage that governs the performance sensitivity in the downstream stage across different hyperparameter settings: surprisingly, a weight decay parameter in SSL plays a crucial role in determining the transferability of features. All SSL backbones exhibit considerable stability and robustness in their performance across diverse hyperparameter configurations when trained with a suitable weight decay parameter during pretraining. The degree of robustness is comparable to that of a supervised pretrained backbone, which is considered to have rich discriminative features. However, it is worth mentioning that determining the appropriate weight decay value during upstream pretraining can be challenging since this value does not have a significant impact on the discriminability evaluated by LP or $k$-NN.

Our main contributions can be summarized as follows:
\begin{itemize}
  \item Based on our extensive experiments, we find that the performance of existing SSL methods is highly sensitive to the hyperparameters utilized in LP and TL, highlighting the inadequacy of current evaluation schemes in assessing the quality of SSL representations.  
  \item We find that the cause of performance variation in LP is unnormalized inputs, and demonstrate that applying BN can eliminate such variation and lead to a more rigorous discriminability comparison.
  \item We demonstrate that a weight decay parameter in SSL controls the transferability of learned representations, despite having little effect on discriminability. This implies that the current TL protocol needs to be reconsidered, as sweeping a large hyperparameter space hinders a fair assessment of transferability.
\end{itemize}
\section{Related Work}
\label{sec:related}

\paragraph{Self-supervised Learning (SSL).} 
The SSL approach involves formulating a pretext task that is capable of leveraging unlabeled data to generate visual representations. Through training on a pretext task, a model learns to capture the underlying structure of the input data. That is, the goal of SSL methods is to build a model capable of learning meaningful representations of the input data. 

Contrastive learning such as SimCLR~\cite{SimCLR} and MoCo~\cite{MoCo} is one of the most popular techniques. In contrastive learning, differently transformed views of the same original image are considered as positive pairs, while transformed views of different images are considered as negative pairs. There are some attempts to conduct SSL using the teacher-student framework like BYOL~\cite{BYOL} and DINO~\cite{DINO}. These methods typically set the teacher model as a moving average of the student model, as opposed to the conventional teacher-student framework for knowledge distillation, which incorporates a pretrained model as the teacher. MAE~\cite{MAE} is one of the representative methods of the masked image modeling (MIM) approach. In MIM, an image is randomly masked and a model is trained to reconstruct the masked patches of the image. Recently, some advanced works such as iBOT~\cite{iBOT}, Mugs~\cite{Mugs}, and MSN~\cite{MSN} have introduced, focusing on training models with multiple pretext tasks. These works aim to enhance representations by leveraging the advantages of each SSL approach.

\paragraph{Evaluation Protocols for SSL.}
The effectiveness of the learned representations via SSL has been assessed from the viewpoints of discriminability and transferability.

To evaluate discriminability, LP has been widely used across previous works, ranging from early-stage research such as Colorization~\cite{Colorization} and RotNet~\cite{RotNet} to more recent works like DINO, MoCo v3, MAE, iBOT, Mugs, and MSN. It is also employed for hyperparameter tuning in SSL since it is a relatively quick evaluation protocol.
Despite the simplicity of LP, which involves training only a linear classifier, we can easily observe that training configurations of existing SSL methods for LP exhibit a wide range of variations.
For example, fundamental hyperparameters like batch size and learning rate vary widely, and MAE employs BN to minimize the need for learning rate expolaration, while other approaches such as DINO, MoCo v3, and Mugs do not consider BN. Moreover, the class tokens from the last four layers of ViT-S are concatenated for LP in DINO, whereas others utilize only the final output features.
In addition to LP, Caron~\textit{et al.}~\cite{DINO} also utilized $k$-NN for simpler evaluation since it has only a single hyperparameter $k$ and no training is necessary.
Following DINO, iBOT and Mugs also validated the quality of representations using the $k$-NN evaluation scheme. However, based on the experimental results of those methods, it is noticeable that there is a rank mismatch in performance between the $k$-NN and LP evaluation results. 

Prior research has also assessed the transferability of representations based on TL, particularly FT, as an additional evaluation method with various downstream datasets. This is an important aspect since the aim of SSL is to acquire general-purpose visual representations that can be valuable for other unseen images. Regarding the hyperparameter settings for FT, some recent works such as DINO, iBOT, Mugs, and MSN have adopted the default setting of DeiT~\cite{DeiT}. However, the DeiT setting requires an excessive amount of computation during FT, e.g., 1,000 training epochs on CIFAR-10/100~\cite{CIFAR} and Stanford Cars~\cite{Car}, and employs very strong regularization techniques (e.g. Mixup~\cite{Mixup}, Cutmix~\cite{Cutmix}, and Random Erasing~\cite{ErasingProb}). On the other hand, MoCo v3 and MAE perform only 100 epochs of training and apply a higher learning rate compared to the DeiT setting. In short, similar to LP, the hyperparameter settings for FT in existing SSL methods exhibit notable variability and tend to be overly tailored to individual downstream datasets. As a result, it becomes challenging to conduct a comprehensive evaluation of whether the learned SSL backbones can serve as general-purpose feature extractors.

To the best of our knowledge, no study has yet highlighted the issues with the current evaluation settings, except for Gwilliam~\textit{et al.}~\cite{Beyond}, who observed that no SSL method emerges as a clear winner under the existing evaluation settings. To facilitate comparison across SSL methods, they analyzed the learned representations using metrics such as uniformity, tolerance, and centered kernel alignment (CKA). Unlike~\cite{Beyond}, our work focuses on identifying the source of performance variations under the existing evaluation protocols rather than comparing SSL methods, and provides recommendations for addressing evaluation issues that require attention.
\begin{table*}[!ht]
    \centering
        \scalebox{0.9}{
        \begin{tabular}{l|c|c|c|ccc|cc}
            \toprule
            & & \textbf{\textit{k}-NN} & \multicolumn{6}{c}{\textbf{LP setting}} \\
            \cline{3-9}
            \textbf{Method} & \textbf{Architecture} & \textbf{\textit{k}=20} & \textbf{Paper} & \textbf{DINO} & \textbf{MoCo v3} & \textbf{MAE} & \textbf{DINO w/BN} & \textbf{MoCo v3 w/BN} \\
            \midrule
            \textbf{SL (DeiT)}~\cite{DeiT} & ViT-S/16 & {79.3} & {79.8} &{79.43} &{78.85} & {79.23} & {79.33} & {78.83} \\
            \textbf{DINO}~\cite{DINO} & ViT-S/16 & {74.4} & {77.0} &{76.86} & {70.53} & {75.91} & {76.27} & {75.26}\\
            \textbf{MoCo v3}~\cite{MoCo_v3} & ViT-S/16 & {68.2} & {73.2} & {47.84} & {73.20} & {72.91} & {73.10} & {72.10}\\
            \textbf{Mugs}~\cite{Mugs} & ViT-S/16 & {75.4} & {78.9} & {77.94} & {72.42} & {77.67} & {77.53} & {76.03}\\
            \textbf{iBOT}~\cite{iBOT} & ViT-S/16 & {74.9} & {77.9} & {77.83} & {71.48} & {76.87} & {77.18} & {75.90}\\
            \textbf{MSN}~\cite{MSN} & ViT-S/16 & {74.8} & {76.9} & {76.63} & {73.68} & {76.31} & {76.76} & {75.56}\\
            \textbf{MAE}~\cite{MAE} & ViT-B/16 & {27.3} & {68.0} & {32.30} & {59.09} & {67.27} & {67.40} & {67.35}\\
            \bottomrule
        \end{tabular}}
     \vspace{5pt}
    \caption{\textbf{${k}$-NN and LP classification results on ImageNet under cross-settings.} We report top-1 accuracy for ${k}$-NN and LP evaluations on ImageNet validation dataset, considering BN. DeiT shows the most stable results. MoCo v3 and MAE show the most unstable results. Applying input normalization using BN resolves the performance instability issue.}
    \label{table:k-nn/lp}
\end{table*}

\section{Discriminability: Normalization Matters}
\label{sec:method1}

In this section, we perform an empirical evaluation of \textit{k}-NN and LP on various SSL methods with cross-setting experiments to investigate how evaluation results differ according to hyperparameter settings and to determine which factors in the training settings are critical for reliable performance evaluation.
\paragraph{Experimental settings.}
We conducted experiments on six SSL methods: DINO~\cite{DINO}, MoCo v3~\cite{MoCo_v3}, iBoT~\cite{iBOT}, Mugs~\cite{Mugs}, MAE~\cite{MAE},  and MSN~\cite{MSN}, using the Vision Transformer architecture (ViT)~\cite{ViT}. 
To ensure a fair comparison, we used ImageNet-1K pretrained official checkpoints of each method. Specifically, we utilized the ViT-S/16 model\footnote{ViT-S of MoCo v3 uses 12 attention heads, while others have 6 heads.}, and for MAE, we used the ViT-B model since the ViT-S checkpoint is currently unreleased. We also compared DeiT~\cite{DeiT} as a representative model for supervised learning. Since the DeiT model is already trained for ImageNet-1K classification with class labels, the LP performance of DeiT could be regarded as an upper-bound reference.

To compare the performance across LP settings of these SSL checkpoints, we selected three hyperparameter settings; DINO, MoCo v3, and MAE settings. These settings were selected as representative training configurations for contrastive learning, teacher-student approaches, and masked image modeling, respectively, considering differences in several hyperparameters such as batch size, learning rate, etc.
For the comparison of \textit{k}-NN performance, we set \textit{k} to 20 because we observed little performance variation when sweeping through the values of $\textit{k} \in \{5, 10, 20, 50, 100\}$. More details on the datasets used, LP experiment settings, and \textit{k}-NN experimental results are provided in the supplementary material (Sections~\ref{appendix/datasets}, \ref{appendix/LP_setting}, and \ref{appendix/kNN_result}, respectively).

\subsection{LP Results with Cross-settings}
Table~\ref{table:k-nn/lp} shows the results of each SSL method evaluated using \textit{k}-NN and LP across DINO, MoCo v3, and MAE settings. Unsurprisingly, the supervised model (DeiT) shows the most stable performance across all evaluations, including \textit{k}-NN and linear evaluation with different settings. On the other hand, SSL models show large differences between evaluations across different LP settings. When comparing the performance of each model based on \textit{k}-NN and LP classification accuracy, there are cases where the rankings of \textit{k}-NN and LP do not align. For example, if we rank \textit{k}-NN performance from 1 to 7, the LP performance of each model with the MoCo v3 setting shows a ranking of 1-4-6-2-3-5-7.
Based on these evaluation results, the selection of the superior model varies depending on the hyperparameter settings used for LP evaluation. However, it can be seen that the MAE setting shows similar performance across multiple SSL checkpoints including MAE and MoCo v3 models, and the \textit{k}-NN accuracy rank aligns with that of LP under the MAE setting. Thus, even if there exists hyperparameter sensitivity in SSL models themselves, we can measure the performance consistently by employing the MAE setting for LP.

\begin{figure}[t]
    \centering
    \subfigure[w/o BN]{
    \begin{minipage}[t]{0.465\linewidth}
      \centering
      \includegraphics[width=\textwidth]{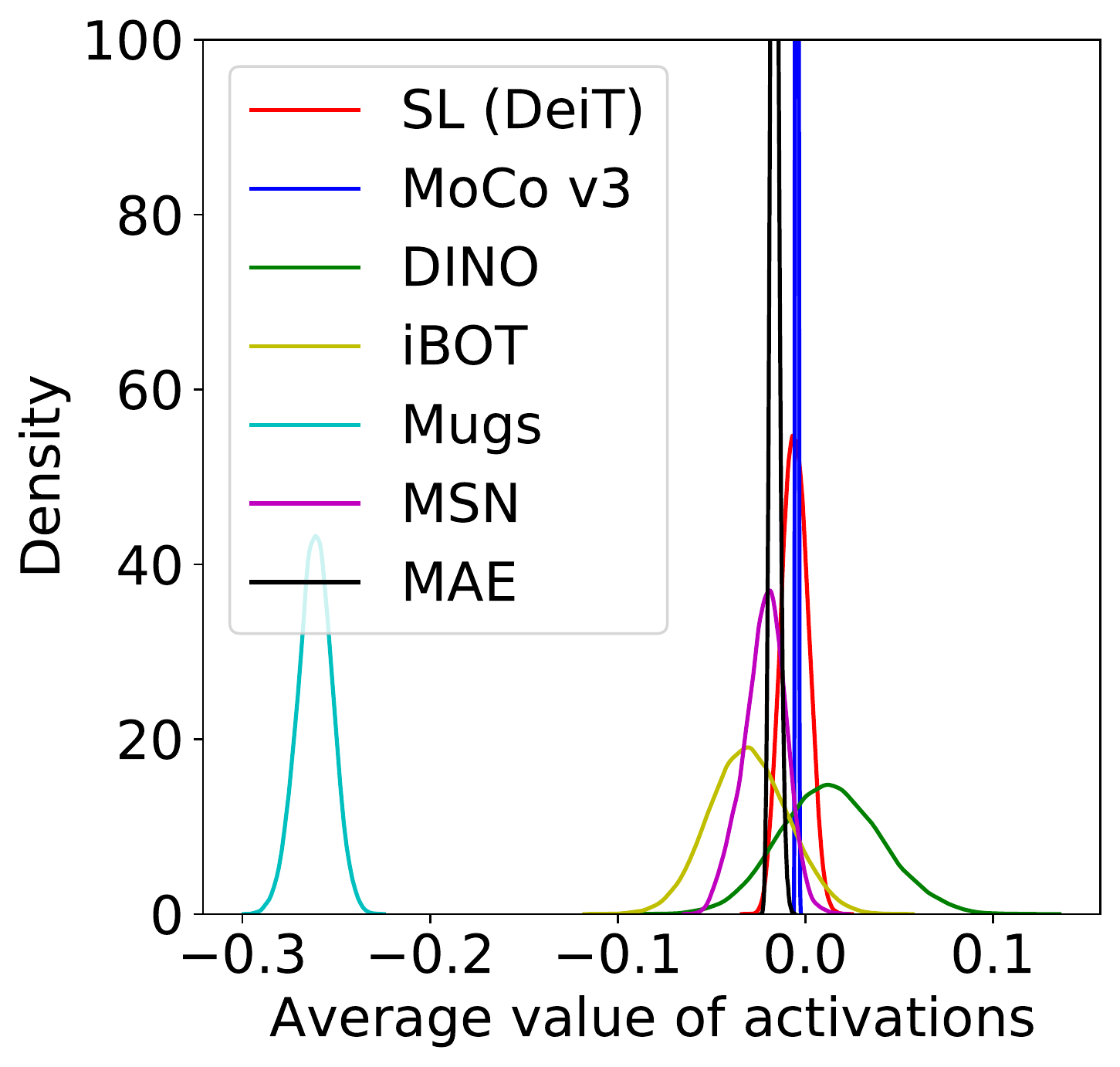}
    \end{minipage}
    }
     \subfigure[w/ BN]{
     \begin{minipage}[t]{0.45\linewidth}
       \centering
       \includegraphics[width=\textwidth]{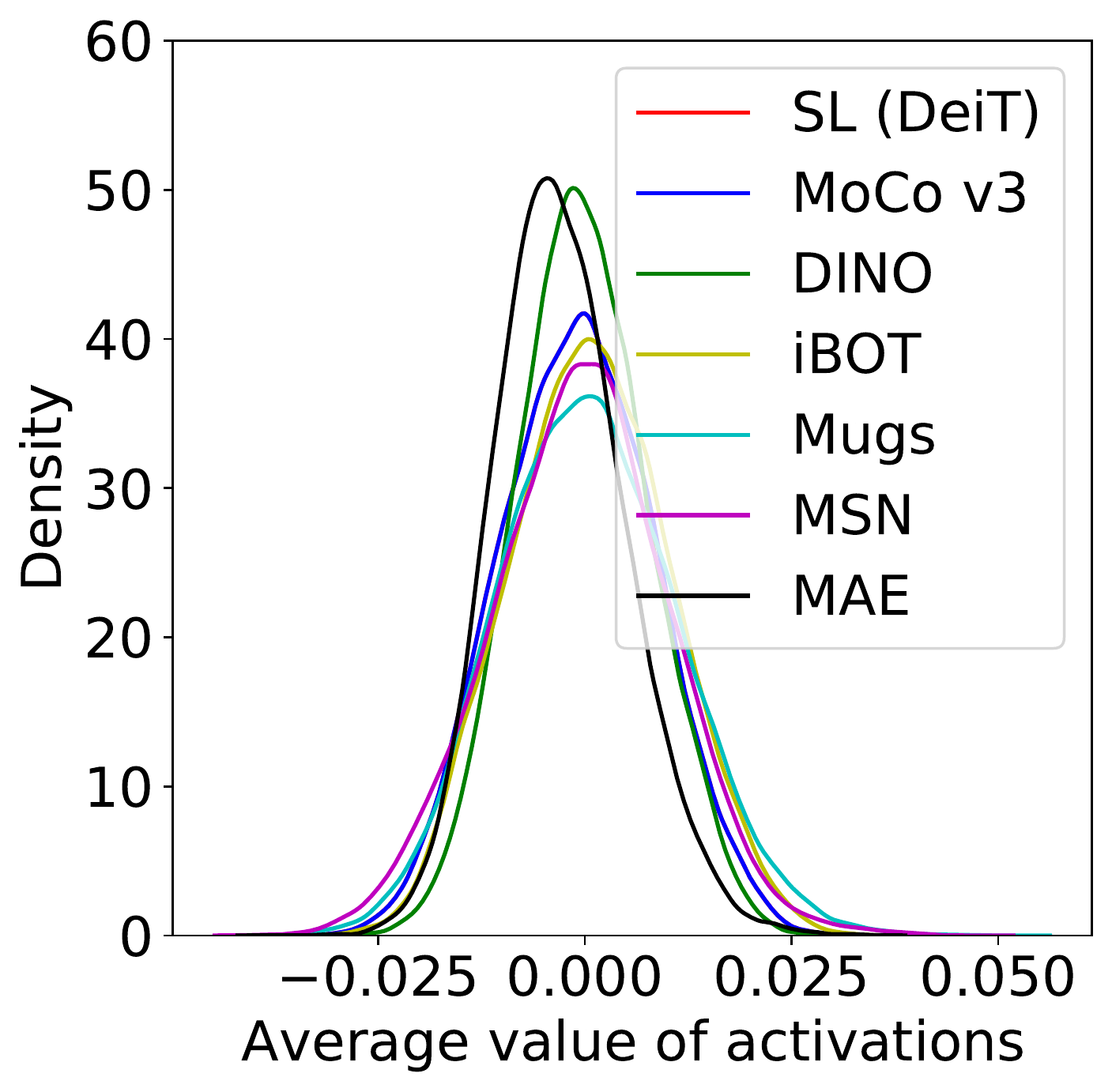}
     \end{minipage}
     }
     
     \caption{\textbf{The activation distribution from each checkpoint with and without BN.} 
            We visualize the feature distribution from each model on the ImageNet validation dataset. Variations in distributions can affect the hyperparameter sensitivity, and using BN can standardize distributions across different models. 
            }
     \label{fig:activations/lpbn}
\end{figure}

\subsection{Input Normalization Effect on LP}
Then, what is the underlying factor that contributes to the stable performance of the MAE setting? The MAE setting has one distinct feature: the application of BN before the linear classifier, which sets it apart from other settings.
As shown in Figure~\ref{fig:activations/lpbn}, representations obtained from each SSL checkpoint show quite different distributions. For example, the features from the Mugs model have notably lower average values compared to those from other checkpoints, while the features produced by the MAE and MoCo v3 models show relatively wider distributions (i.e., high variance). In LP, represented as $\mathbf{y=wx+b}$, differences in mean values can be adjusted by changing the bias vector $\mathbf{b}$, but differences in variance require adjusting the scale of each weight in $\mathbf{w}$, which can potentially affect hyperparameter sensitivity. However, when BN is applied, the features extracted from each model show substantially similar distributions, which can reduce the sensitivity.

\begin{figure}[t]
    \centering
        \subfigure[\textit{k}-NN / LP (w/o BN)]{
    \begin{minipage}[t]{0.465\linewidth}
      \centering
      \includegraphics[width=\textwidth]{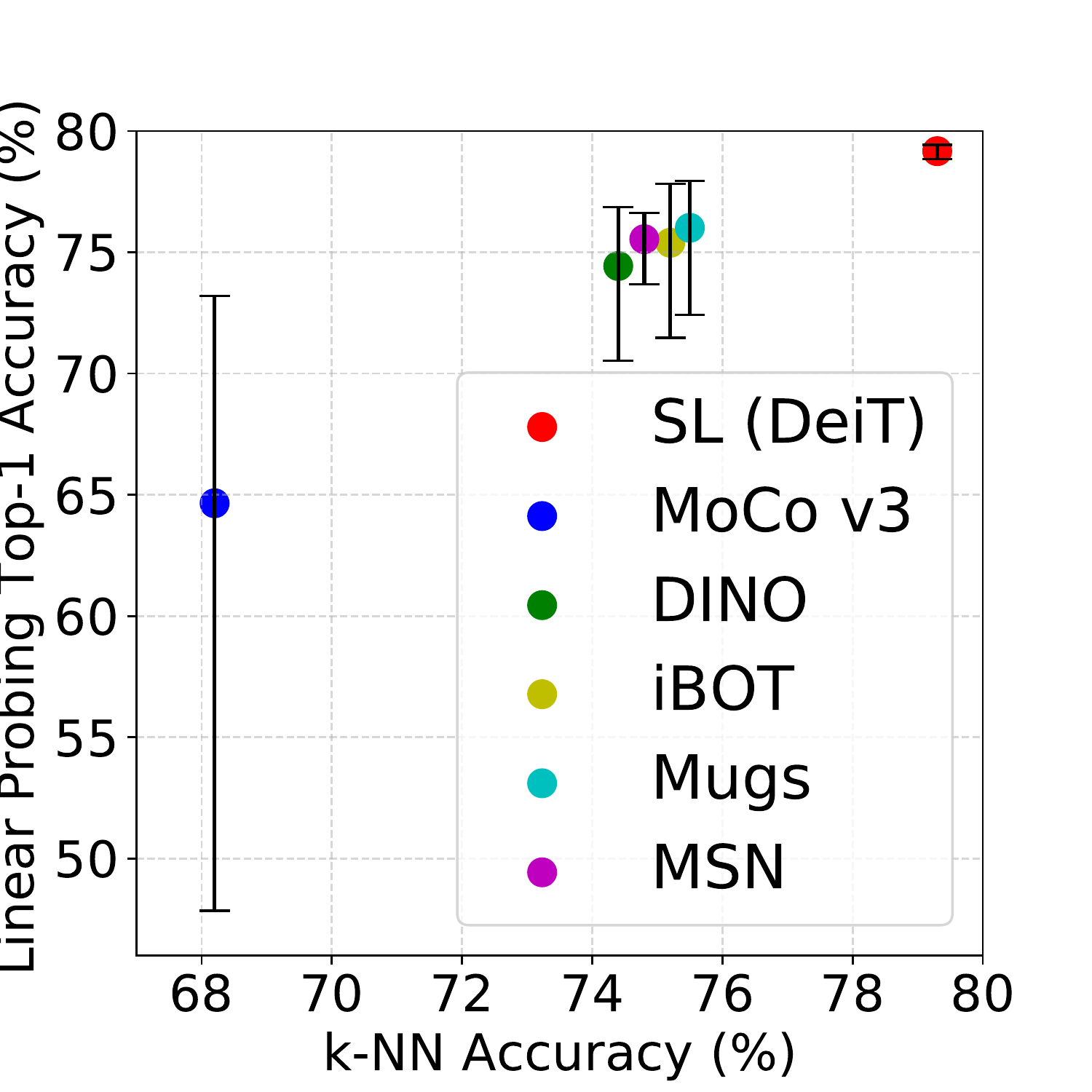}
    \end{minipage}
  }
         \subfigure[\textit{k}-NN / LP (w/ BN)]{
    \begin{minipage}[t]{0.465\linewidth}
      \centering
      \includegraphics[width=\textwidth]{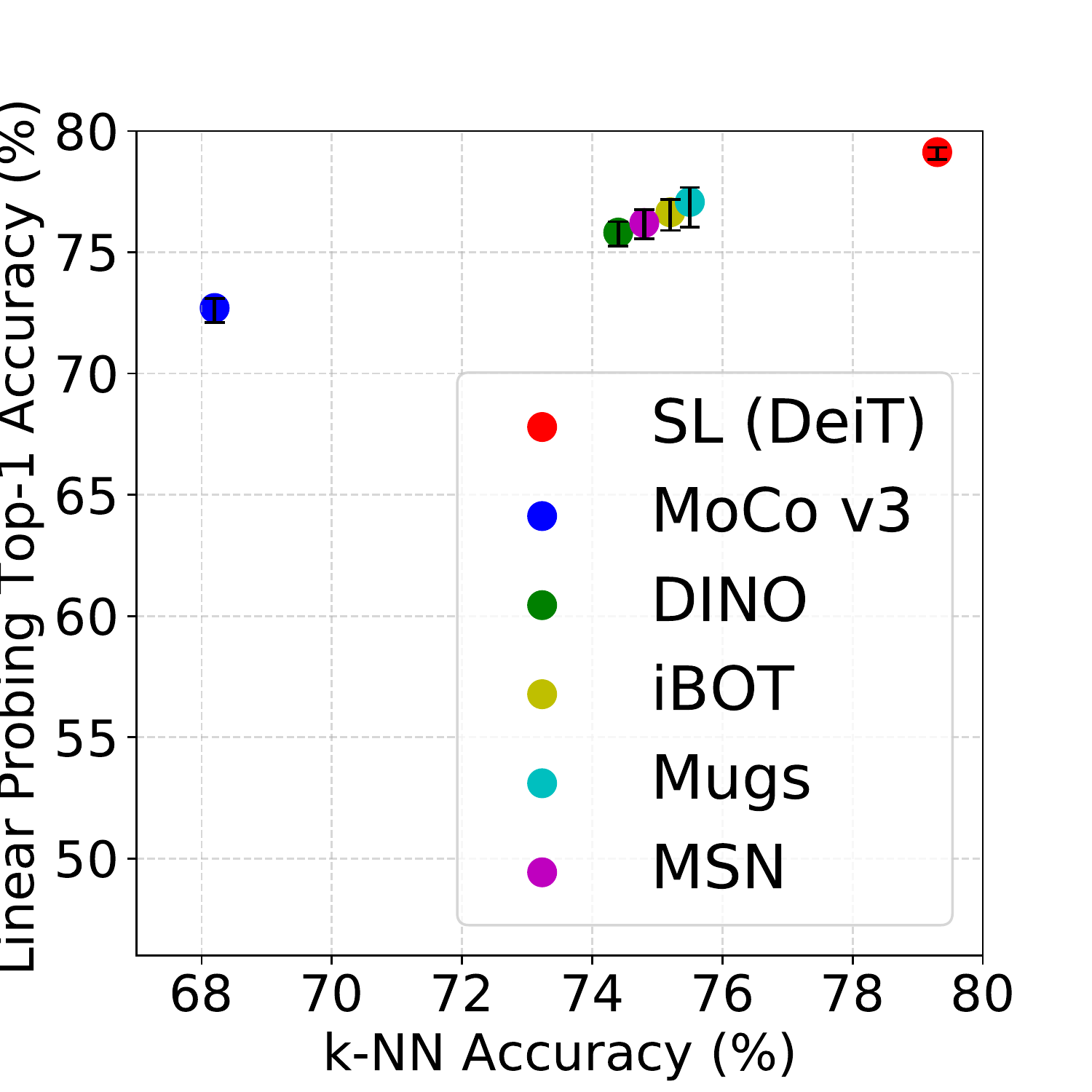}
    \end{minipage}
  }
     \caption{\textbf{ImageNet validation top-1 accuracy of \textit{k}-NN and LP with and without BN.} When representing multiple accuracy values for LP settings, we present the median value using a point and the range of values using error bars to show the minimum and maximum values.}
     \label{fig:knn/lpbn}
\end{figure}

To observe the performance stabilization effect of input normalization on the DINO and MoCo v3 settings, we apply BN, while keeping the learnable scale and shift parameters in BN fixed to 1 and 0, respectively. As shown in Table~\ref{table:k-nn/lp}, applying BN to the DINO and MoCo v3 settings can improve overall performance and robustness. When comparing the performance of each model under different settings, it can be observed that there is a significant difference of 3--26\% without BN, but after applying BN, the difference is reduced to around 1\%. Despite the difference in various factors such as batch size, optimizer, and learning rate, similar performance is observed across all settings.
By looking at \textit{k}-NN versus LP plot in Figure~\ref{fig:knn/lpbn}, we can clearly identify how BN influences the performance difference among different LP settings.
The use of BN helps to stabilize the hyperparameter sensitivity of SSL checkpoints, and also aligns the ranking of \textit{k}-NN with that of LP.

In summary, LP is highly sensitive to hyperparameter settings when evaluating the discriminability of learned representations, making it difficult to draw meaningful conclusions. However, the use of BN can significantly reduce this instability. Based on these findings, we highly recommend the use of BN as a standard evaluation protocol for discriminability, applied before the classification layer.

\section{Transferability: A Hidden Parameter Exists}
\label{sec:method2}

\begin{table}[t]
    \centering
    \resizebox{1.0\linewidth}{!}{%
    \begin{tabular}{l|l|c|ccc|c}
    \toprule
     &  & \multicolumn{5}{c}{\textbf{TL setting}} \\
    \cline{3-7}
    \textbf{Dataset} & \textbf{Method} & \textbf{Paper} & \textbf{DINO} & \textbf{MoCo v3} & \textbf{Short} & \textbf{Gap} \\
    \midrule
    \multirow{6}{*}{\textbf{CIFAR-100}} & \textbf{SL (DeiT)} & 89.5 & \textbf{90.02} & \textbf{88.14} & \textbf{85.89} & \textbf{4.13} \\
    & \textbf{DINO} & 90.5 & 90.21 & 72.46 & {57.40} & {32.81} \\
    & \textbf{MoCo v3} & - & 88.79 & 89.06 & 80.88 & 8.18 \\
    & \textbf{Mugs} & 91.8 & 91.36 & 88.06 & 86.03 & 5.33 \\
    & \textbf{iBOT} & 90.7 & \textit{\underline{90.60}} & \textit{\underline{68.73}} & \textit{\underline{51.49}} & \textit{\underline{39.11}} \\
    & \textbf{MSN} & 90.5 & 90.31 & 86.67 & 79.21 & 11.10 \\
    \midrule
    \multirow{6}{*}{\textbf{Flowers-102}} & \textbf{SL (DeiT)} & 98.2 & \textbf{97.04} & \textbf{96.99} & \textbf{94.53} & \textbf{2.51} \\
    & \textbf{DINO} & 98.5 & 98.24 & {37.42} & 86.61 & {60.82} \\
    & \textbf{MoCo v3} & - & 83.73 & 94.67 & 86.61 & 10.94 \\
    & \textbf{Mugs} & 98.8 & 97.97 & 96.06 & 94.58 & 3.39 \\
    & \textbf{iBOT} & 98.6 & \textit{\underline{98.55}} & \textit{\underline{31.57}} &\textit{\underline{55.05}} & \textit{\underline{66.98}} \\
    & \textbf{MSN} & - & 98.36 & 94.80 & 92.29 & 6.07 \\
    \midrule
    \multirow{6}{*}{\textbf{Stanford Cars}} & \textbf{SL (DeiT)} & 92.1 & \textbf{88.76} & \textbf{88.46} & \textbf{80.27} & \textbf{8.49} \\
    & \textbf{DINO} & 93.0 & 92.24 & {10.97} & {8.69} & {83.55} \\
    & \textbf{MoCo v3} & - & {49.27} & 89.43 & {41.49} & {47.94} \\
    & \textbf{Mugs} & 93.9 & 91.11 & 89.45 & 80.07 & 11.04 \\
    & \textbf{iBOT} & 94.0 & \textit{\underline{92.35}} & \textit{\underline{11.62}} & \textit{\underline{6.11}} & \textit{\underline{86.24}} \\
    & \textbf{MSN} & - & 92.71 & 80.24 & {7.98} & {84.73} \\
    \midrule
    \multirow{6}{*}{\textbf{iNaturalist19}} & \textbf{SL (DeiT)} & 76.6 & \textbf{76.34} & \textbf{74.16} & \textbf{62.64} & \textbf{13.70} \\
    & \textbf{DINO} & 78.2 & \textit{\underline{78.75}} & \textit{\underline{56.04}} & \textit{\underline{36.92}} & \textit{\underline{41.83}} \\
    & \textbf{MoCo v3} & - & 76.01 & 72.05 & {58.65} & 17.36 \\
    & \textbf{Mugs} & 79.8 & 78.65 & 75.38 & 63.77 & 14.88 \\
    & \textbf{iBOT} & 78.5 & 77.89 & {53.00} & {37.18} & {40.71} \\
    & \textbf{MSN} & 78.1 & 78.22 & 73.10 & {47.72} & {30.50} \\ 
    \bottomrule
\end{tabular}}
     \vspace{5pt}
    \caption{\textbf{TL results on downstream datasets.} We report top-1 accuracy on various downstream validation datasets. We use the same pretrained checkpoints used in Table~\ref{table:k-nn/lp}. Supervised backbone shows stable results regardless of hyperparameter settings, but some SSL backbones show unstable outcomes. ``Gap'' refers to the difference between the maximum and minimum values among three settings.
    Models with the lowest and highest ``Gap'' for each dataset are shown in bold and underlined, respectively.
    }
\label{table:FT_result}
\end{table}

In addition to LP and \textit{k}-NN, TL is one of the popular evaluation protocols for SSL representations. While SSL aims to create models with rich representations of the upstream data domain, it is more important to focus on TL rather than LP and \textit{k}-NN when evaluating the practical usability of these representations in real-world applications. However, similar to the findings in the previous section, TL hyperparameter settings and their performances vary significantly across SSL methods. In this section, we conduct extensive experiments to investigate this phenomenon and explore which properties of a pretrained model mainly affect its transferability.

\paragraph{Experimental settings.}
We compared the same SSL checkpoints as in Section~\ref{sec:method1} except MAE which showed low LP performance. The datasets used for TL are CIFAR-100~\cite{CIFAR}, Oxford Flowers 102~\cite{Flower}, Stanford Cars~\cite{Car}, and iNaturalist 2019~\cite{iNat}. Unlike in the evaluation of discriminability where the target domain for training and evaluation is matched for the supervised model, in the case of TL, the target domain is different for both supervised and SSL models. Therefore, the supervised model does not set a specific upper bound for the SSL models' performance.

For cross-setting experiments on TL, by considering the learning rate, training epochs, and regularization, the DINO setting was selected as the representative of the \textit{slow-long-strong} setting (\textit{slow} learning by a low learning rate, \textit{long} training epochs, and \textit{strong} regularization), and the MoCo v3 setting was selected as the \textit{fast-short-strong} one. Additionally, we configured the \textit{fast-short-weak} setting named ``Short'' since long training with strong regularization (such as the \textit{slow-long-strong} DINO setting) could lead to high performance regardless of the transferability of learned representations. More details on datasets and experimental settings are available in the supplementary material (Sections~\ref{appendix/datasets} and \ref{appendix/TL_setting}, respectively)

\begin{table*}[t]
    \centering
    \resizebox{0.88\linewidth}{!}{%
    \begin{tabular}{l|l|c|ccc|ccc|ccc}
    \toprule
    \textbf{} & \multicolumn{1}{c|}{\textbf{SSL setting (IN)}} & \textbf{LP setting (IN)} & \multicolumn{3}{c|}{\textbf{TL setting (Cars)}} & \multicolumn{3}{c|}{\textbf{TL setting (Flowers)}} & \multicolumn{3}{c}{\textbf{TL setting (CIFAR)}} \\
    \cline{2-12}
    \textbf{Method} & \textbf{Weight decay} & \textbf{MoCo v3 w/ BN} & \textbf{DINO} & \textbf{MoCo v3} & \textbf{Short} & \textbf{DINO} & \textbf{MoCo v3} & \textbf{Short} & \textbf{DINO} & \textbf{MoCo v3} & \textbf{Short} \\
    \midrule
    \multirow{4}{*}{\textbf{Mugs}} & 0.04 $\,\to\,$ 0.4 & 73.98 & \textit{\underline{91.48}} & \textit{\underline{73.85}} & \textit{\underline{44.98}} & \textit{\underline{97.17}} & \textit{\underline{44.71}} & \textit{\underline{91.75}} & \textit{\underline{89.49}} & \textit{\underline{86.80}} & \textit{\underline{83.80}} \\
    & 0.04 $\,\to\,$  0.2* & 74.46 & 91.21 & 89.00 & 74.04 & \textbf{97.11} & \textbf{96.78} & \textbf{95.04} & 89.21 & 88.20 & 86.20 \\
    & 0.04 & 74.00 & \textbf{87.86} & \textbf{89.58} & \textbf{79.27} & 95.30 & 97.37 & 93.86 & \textbf{89.08} & \textbf{89.31} & \textbf{88.79} \\
    & 0 & 70.56 & {46.44} & 84.99 & 68.91 & 64.63 & 80.22 & 89.19 & 84.46 & 88.24 & 86.75 \\
    \midrule
    \multirow{4}{*}{\textbf{DINO}} & 0.04 $\,\to\,$ 0.4* & 72.77 & \textit{\underline{91.26}} & \textit{\underline{23.67}} & \textit{\underline{19.92}} & 97.02 & 88.32 & 93.02 & \textit{\underline{88.62}} & \textit{\underline{87.20}} & \textit{\underline{79.66}} \\
    & 0.04 $\,\to\,$  0.2 & 73.09 & 90.80 & 86.31 & 73.39 & \textbf{97.06} & \textbf{96.11} & \textbf{94.98} & 88.39 & 87.74 & 85.49 \\
    & 0.04 & 72.39 & \textbf{85.50} & \textbf{86.37} & \textbf{77.45} & 95.28 & 96.89 & 94.34 & \textbf{87.68} & \textbf{88.65} & \textbf{87.36} \\
    & 0 & 67.65 & {58.08} & 85.14 & 69.65 & \textit{\underline{78.26}} & \textit{\underline{89.01}} & \textit{\underline{91.35}} & 82.55 & 86.45 & 85.14 \\
    \midrule
    \multirow{5}{*}{\textbf{MoCo v3}} & 0.04 $\,\to\,$ 0.4 & 61.35 & {68.00} & {9.76} & {4.28} & \textit{\underline{88.36}} & \textit{\underline{12.23}} & \textit{\underline{36.55}} & \textit{\underline{84.06}} & \textit{\underline{64.23}} & \textit{\underline{57.32}} \\
    & 0.04 $\,\to\,$ 0.2 & 64.90 & 69.53 & 49.61 & 5.50 & 89.80 & {22.36} & 77.86 & 84.75 & 82.97 & 73.02 \\
    & 0.1* & 65.37 & \textit{\underline{68.01}} & \textit{\underline{85.35}} & \textit{\underline{8.92}} & 87.41 & 84.53 & 90.67 & 85.34 & 85.56 & 81.48 \\
    & 0.04 & 65.17 & \textbf{75.58} & \textbf{86.94} & \textbf{73.25} & \textbf{88.53} & \textbf{92.57} & \textbf{91.70} & \textbf{85.17} & \textbf{86.92} & \textbf{84.93} \\
    & 0 & 62.96 & 65.84 & 84.69 & 69.59 & 89.92 & 93.46 & 89.06 & 84.66 & 86.79 & 84.68 \\
    \midrule
    \multirow{4}{*}{\textbf{MSN}} & 0.04 $\,\to\,$ 0.4* & 67.48 & \textit{\underline{88.72}} & \textit{\underline{78.15}} & \textit{\underline{54.49}} & 96.31 & 94.81 & 93.43 & 86.74 & 86.55 & 83.38 \\
    & 0.04 $\,\to\,$ 0.2 & 66.60 & 85.97 & 81.17 & 70.78 & \textbf{95.30} & \textbf{95.27} & \textbf{93.71} & \textbf{86.49} & \textbf{86.98} & \textbf{85.05} \\
    & 0.04 & 65.19 & \textbf{78.50} & \textbf{81.41} & \textbf{67.55} & 92.50 & 94.75 & 91.68 & 84.74 & 86.97 & 84.98 \\
    & 0 & 63.96 & 64.35 & 78.85 & 62.75 & \textit{\underline{87.69}} & \textit{\underline{90.97}} & \textit{\underline{90.99}} & \textit{\underline{83.64}} & \textit{\underline{87.06}} & \textit{\underline{84.62}} \\
    \midrule
    \multirow{4}{*}{\textbf{iBOT}} & 0.04 $\,\to\,$ 0.4* & 74.25 & \textit{\underline{91.98}} & \textit{\underline{55.66}} & \textit{\underline{21.77}} & \textit{\underline{97.40}} & \textit{\underline{39.45}} & \textit{\underline{92.30}} & \textit{\underline{89.42}} & \textit{\underline{66.73}} & \textit{\underline{76.49}} \\
    & 0.04 $\,\to\,$ 0.2 & 74.77 & 91.39 & 87.86 & 71.43 & 97.33 & 95.85 & 95.11 & 89.20 & 87.80 & 85.91 \\
    & 0.04 & 74.23 & \textbf{87.80} & \textbf{89.52} & \textbf{81.09} & \textbf{95.90} & \textbf{97.43} & \textbf{95.30} & \textbf{88.92} & \textbf{89.38} & \textbf{88.79} \\
    & 0 & 70.86 & 61.78 & 87.60 & 73.63 & 81.31 & 91.48 & 92.32 & 85.63 & 88.50 & 87.87 \\
    \bottomrule
    \end{tabular}}
     \vspace{5pt}        
    \caption{\textbf{Analysis of weight decay during upstream pretraining.} We trained ViT-S/16 using each SSL method for 100 epochs with varying weight decay values, and report LP accuracy on ImageNet (IN) as well as TL classification results on Stanford Cars (Cars), Flower-102 (Flowers), and CIFAR-100 (CIFAR). The symbol `*' denotes the default weight decay for each method used in the literature. The weight decay during SSL considerably affects the stability of TL performance whereas LP performance remains robust to changes in weight decay values. Considering the min-max difference among three settings, the most stable results are shown in bold, and the most unstable results are underlined.}
\label{weight_decay_ablation}
\end{table*}

\subsection{TL Results with Cross-settings}
Table~\ref{table:FT_result} shows the performance of different SSL checkpoints on four benchmark datasets across various TL settings. It is challenging to identify a single best-performing SSL checkpoint because the model that achieves the highest performance varies across datasets. Notably, there is a significant performance gap of up to 86.24\%, when examining the TL performance across multiple hyperparameter settings. Furthermore, unlike the LP's MAE setting, there is no preferred single TL setting.

However, surprisingly, the SL (DeiT) model consistently shows stable and high performance unlike SSL models, with a performance gap of 2.51--13.70\%. The inclusion of label supervision during training is likely to have helped the SL (DeiT) model to learn more discriminative representations, leading to increased robustness across various TL settings, particularly since the downstream task is also a classification task. Out of the various SSL models compared, only the Mugs model has achieved a similar level of stability comparable to the SL model. From this perspective, we believe that hyperparameter robustness should also be considered when comparing the performance of SSL models with that of SL models. 

Based on our experimental results, we have confirmed that TL performance variations stem from the inherent nature of SSL checkpoints including model architectures and SSL training recipes rather than TL hyperparameter settings. The Mugs model learns three tasks simultaneously: instance, local-group, and group discrimination. From the ablation study on Mugs, we verified that neither architectural components nor losses contribute to the robust TL performance of the Mugs checkpoint (see the ablation results in the supplementary material, Section~\ref{mugs_abl}). After careful investigation, we conclude that the weight decay value during upstream pretraining is a key factor that controls the transferability of SSL representations.

\subsection{Weight Decay Governs Transferability}
Weight decay is typically used for regularization. If the weight decay value is too small, the regularization effect is weak, while if it is too large, it can hinder training. The impact of SSL weight decay on training can be explained in similar but different ways. It is known that, in non-contrastive SSL based on two augmented views, SSL models learn both nuisance features created by the augmentation process and invariant features based on image content~\cite{DirectCopy}. If the weight decay value is too small, the SSL model may learn nuisance features made by augmentation, leading to overfitting on the upstream data. On the other hand, if the weight decay value is too large, the SSL model may suffer from feature collapse as it fails to properly learn invariant features. The authors of~\cite{DirectCopy} analyzed the impact of weight decay on upstream performance (i.e., discriminability), but not TL performance on downstream datasets.

Learning invariant features of the upstream data can be helpful from the perspective of upstream evaluation, but it may not necessarily be useful from the perspective of TL, as low- and mid-level statistical features are important for TL on various downstream datasets rather than high-level semantic features of the upstream data. Therefore, it is necessary to learn more comprehensive and rich general-purpose representations, rather than just the invariant features of image content. 

\begin{figure}[t]
\vskip 0.2in
\begin{center}
\includegraphics[width=.6\columnwidth]{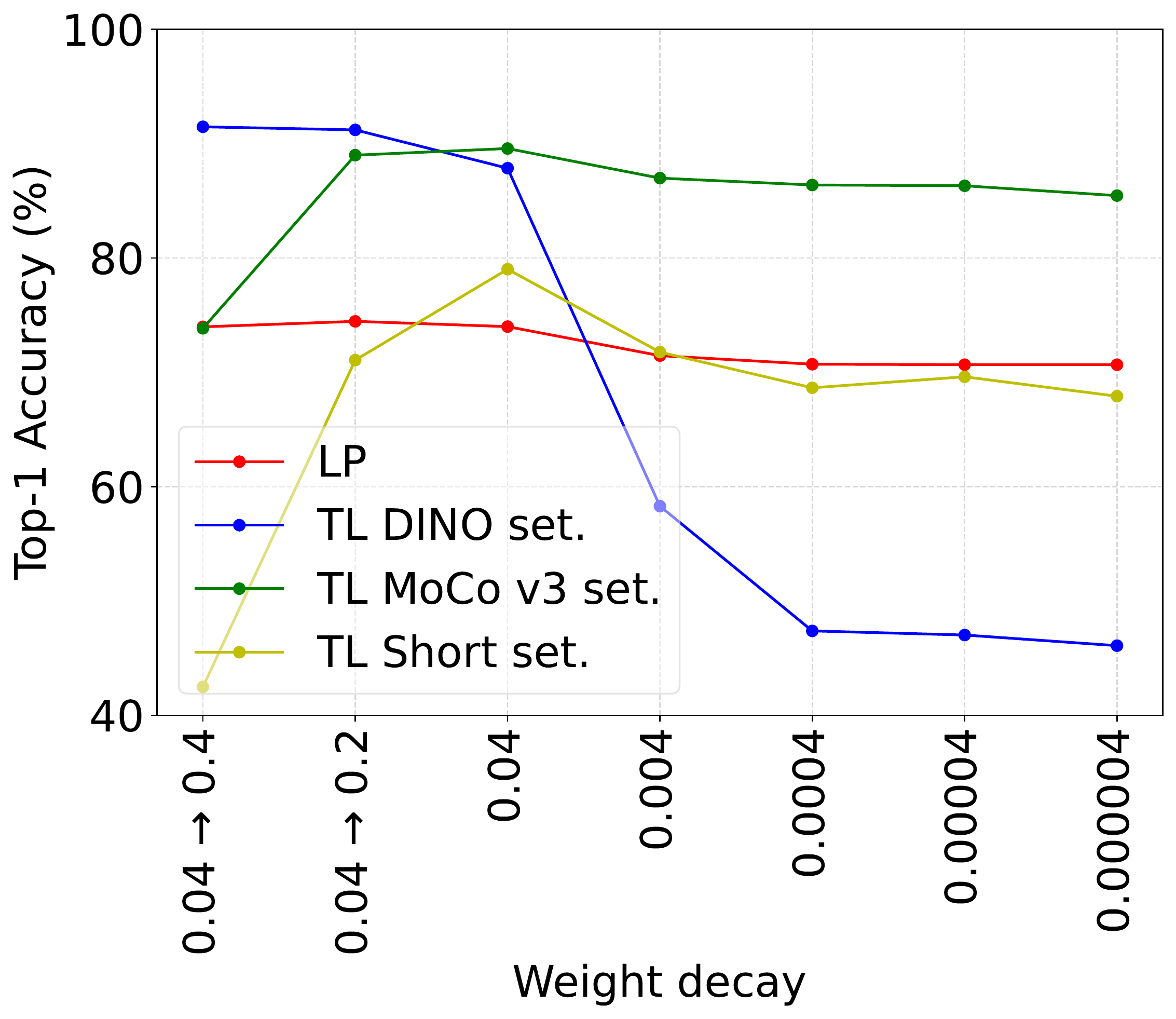}
\caption{
\textbf{Additional analysis on Mugs according to SSL weight decay.} LP shows similar performance even with different weight decay values, but TL performance varies.
}
\label{fig:mugs_wd_abl}
\vskip -0.2in
\end{center}
\vskip -0.2in
\end{figure}

In order to analyze the impact of SSL weight decay on TL, we experimented with several weight decay values, then measured the performance using the same TL settings as in Table~\ref{table:FT_result}. We tested various weight decay values such as 0.04$\,\to\,$0.4 (the default in DINO, MSN, and iBOT), 0.04$\,\to\,$0.2 (the default in Mugs), 0.04, and 0. Following DINO and Mugs, we employed cosine scheduling to increase weight decay values during training. Additionally, we used a weight decay value of 0.1 (the default in MoCo v3) specifically for the MoCo v3 method.

Table~\ref{weight_decay_ablation} summarizes experimental results with various SSL weight decay values on three downstream datasets, CIFAR-100, Oxford Flowers 102, and Stanford Cars. From the perspective of min-max performance differences across TL settings for each dataset, we observe that Mugs and DINO with the SSL weight decay of 0.04$\,\to\,$0.2 exhibit a small difference of 2.07--17.17\% and 2.08--17.41\%, respectively. However, with the increased SSL weight decay of 0.04$\,\to\,$0.4, the performance variations significantly increase to 5.69--52.46\% and 8.70--71.34\%, respectively. This finding is intriguing as it highlights that the performance of TL greatly varies when the SSL weight decay value is slightly changed. This pattern is observed across all SSL methods, for example, iBOT also shows considerable performance variations according to the SSL weight decay: the highest stability is achieved with the SSL weight decay value of 0.04 across all datasets. Interestingly, all models enjoy a high degree of transferability (i.e. stability) when the SSL weight decay value is set to 0.04$\,\to\,$0.2 or 0.04. 

However, such behavior cannot be noticed by comparing LP performance, e.g., iBOT shows LP performances of 74.25\% and 74.23\% for the SSL weight decay of 0.04$\,\to\,$0.4 and 0.04, respectively, while the variation of TL performance is significantly small for the SSL weight decay of 0.04. Figure~\ref{fig:mugs_wd_abl} shows the detailed results for Mugs. From this figure, we can easily observe that LP performances are similar despite changes in SSL weight decay, while TL performances do not. The Mugs model shows the highest stability with the SSL weight decay of 0.04. It should be noted again that, except for the SSL weight decay value, we used the default SSL settings for each method. Considering the fact that the stability/instability caused by SSL weight decay is not limited to a particular dataset or method, but appears consistently across all datasets and models, we can conclude that the performance robustness to TL settings (i.e., transferability) is significantly dominated by the SSL weight decay.

To summarize, evaluating the transferability of SSL representations requires cross-setting experiments rather than just comparing the highest LP and TL accuracy. The instability of SSL representations can only be confirmed through such experiments. Additionally, the weight decay hyperparameter used in upstream pretraining is critical in learning transferable and general-purpose visual representations, and thus, it requires attention when evaluating the transferability of SSL representations.

\subsection{In-depth Analysis of Weight Decay Effect}
\label{sec:analysis}
From the perspective of analyzing how SSL weight decay affects TL sensitivity of SSL models, we conducted a comparative analysis of Mugs and DINO. Firstly, in terms of loss curves, we discuss the changes in training speed resulting from SSL weight decay. Secondly, in terms of loss landscape, we explore the impact of SSL weight decay on the loss landscape and how it can lead to robust TL. Lastly, using CKA, we analyze the impact of SSL weight decay on feature reuse from a transferability perspective. 
We selected the DINO setting (i.e., \textit{slow-long-strong}) as a default since it provides the best-performing models for both Mugs and DINO on the downstream datasets (see Table~\ref{table:FT_result}).

\begin{figure}[t]
    \centering
        \subfigure[Mugs]{
    \begin{minipage}[t]{0.4\linewidth}
      \centering
      \includegraphics[width=\textwidth]{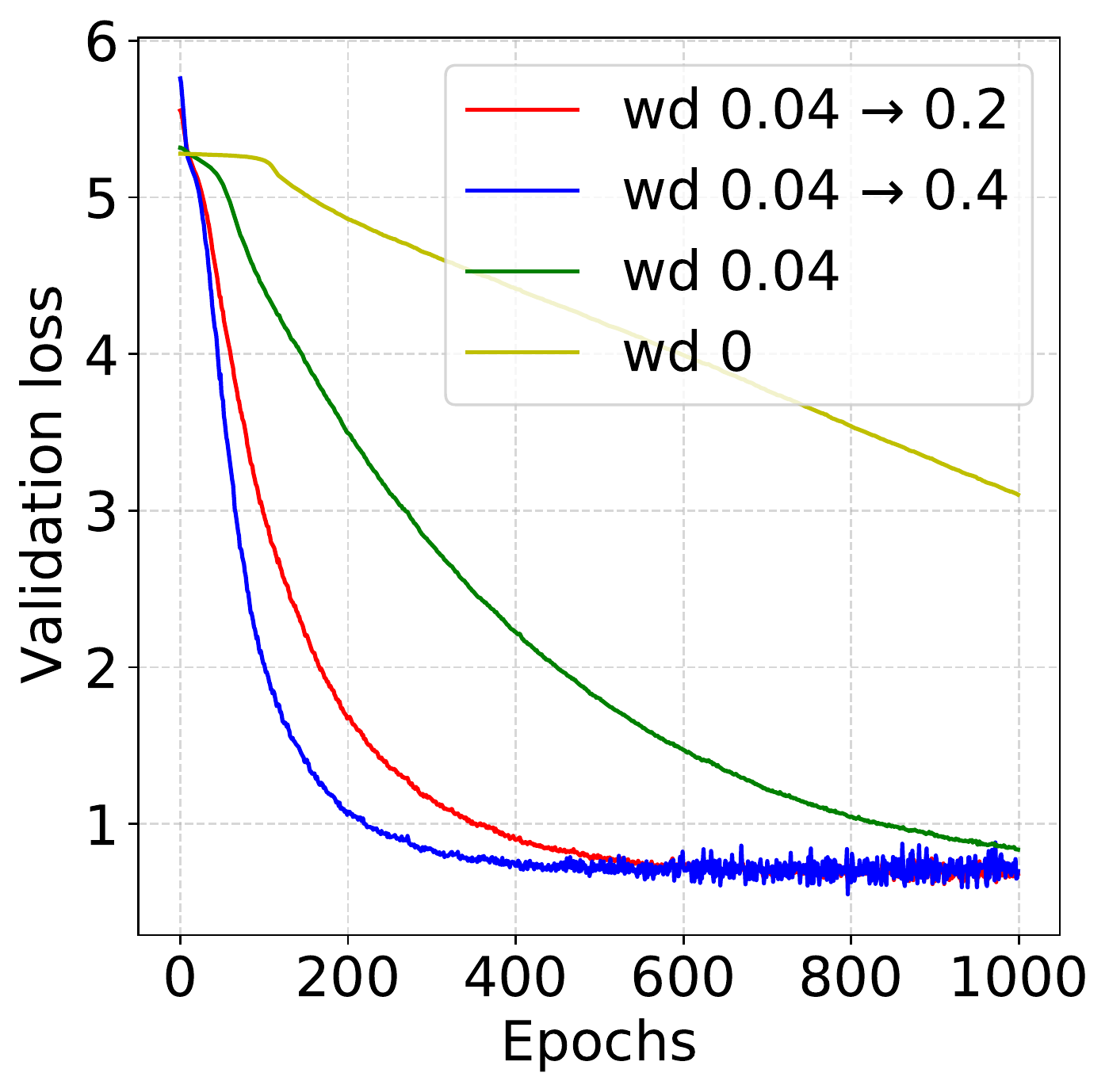}
    \end{minipage}
  }
        \subfigure[DINO]{
    \begin{minipage}[t]{0.4\linewidth}
      \centering
      \includegraphics[width=\textwidth]{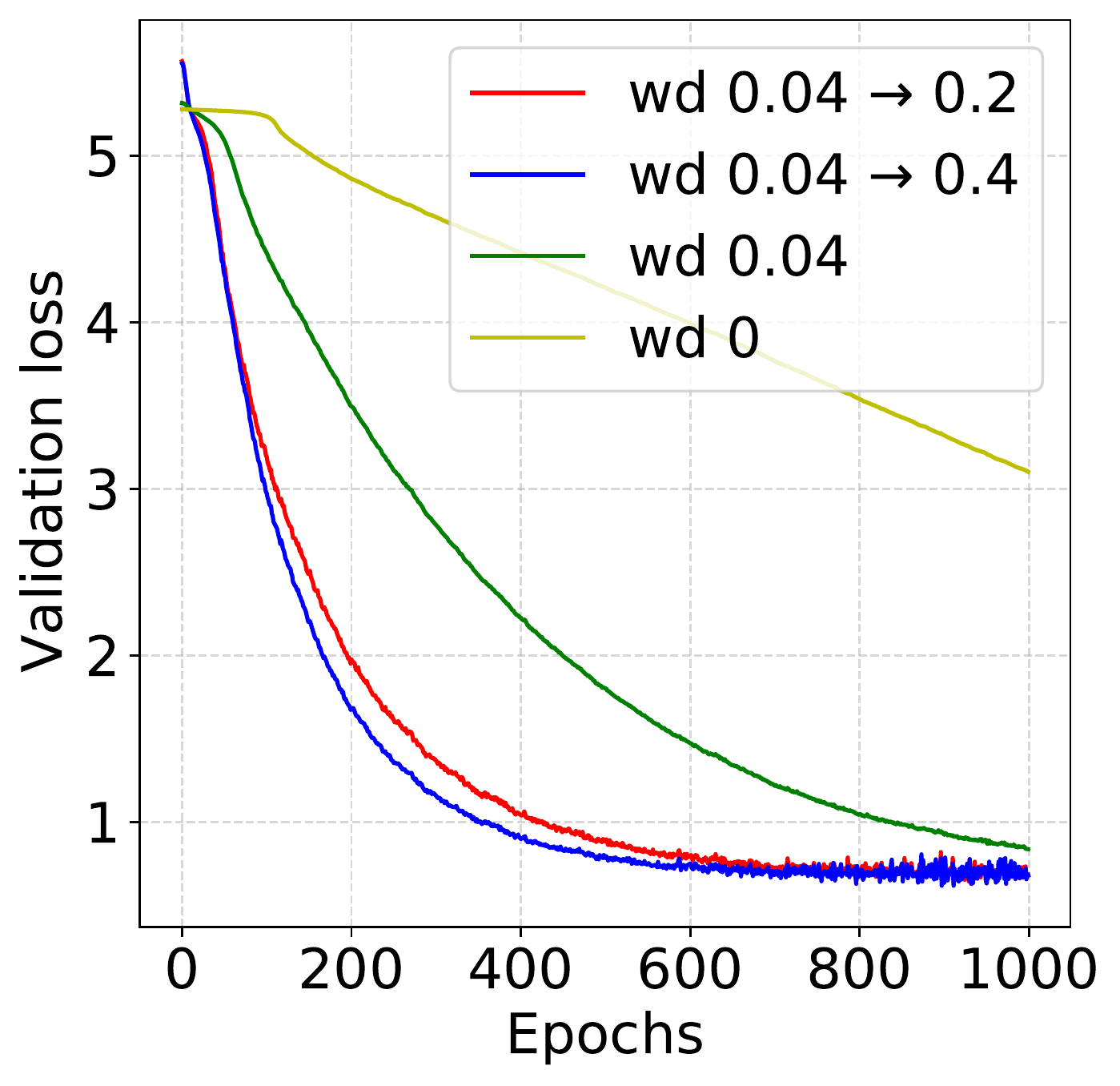}
    \end{minipage}
  }
     \vspace{0pt}
    \caption{
    \textbf{Validation loss curves depending on SSL weight decay.} We use the models trained on Stanford Cars in Table~\ref{weight_decay_ablation}. As the SSL weight decay value decreases, the convergence rate also decreases.
    }
    \label{fig:loss_curve}
\end{figure}

\paragraph{Loss curve.}
\label{loss_curve}
Figure~\ref{fig:loss_curve} shows the changes in the convergence rate of validation losses for Stanford Cars during the TL process of the Mugs and DINO models trained for Table~\ref{weight_decay_ablation}. 
Once again, it should be emphasized that all TL hyperparameters, including weight decay during TL, were fixed in order to isolate the effects of different SSL backbones pretrained with varying SSL weight decay values. 
In Figure~\ref{fig:loss_curve}, it can be noticed that a decrease in SSL weight decay results in a slower learning speed. In the cases of 0.04$\,\to\,$0.2 and 0.04$\,\to\,$0.4, the difference in training speed may not be significant and both achieve similar validation loss at convergence. However, for the cases of 0.04 and 0, it can be observed that the convergence rates are notably low.
It should be noted that such different training behaviors are only visible in TL scenarios, as SSL backbones with different weight decay values show almost identical LP performance except for the case where weight decay is 0 (see Table~\ref{weight_decay_ablation}).
Additionally, faster convergence does not necessarily lead to better transferability. Although Mugs and DINO backbones with the SSL weight decay of 0.04$\,\to\,$0.4 show faster training (and higher accuracy), these backbones do not guarantee a high level of transferability as shown in Table~\ref{weight_decay_ablation}. As a result, validation loss curves alone do not provide sufficient information to estimate the transferability of SSL backbones.

\begin{figure}
\begin{center}
    \makebox[\widthof{\includegraphics[scale=0.25]{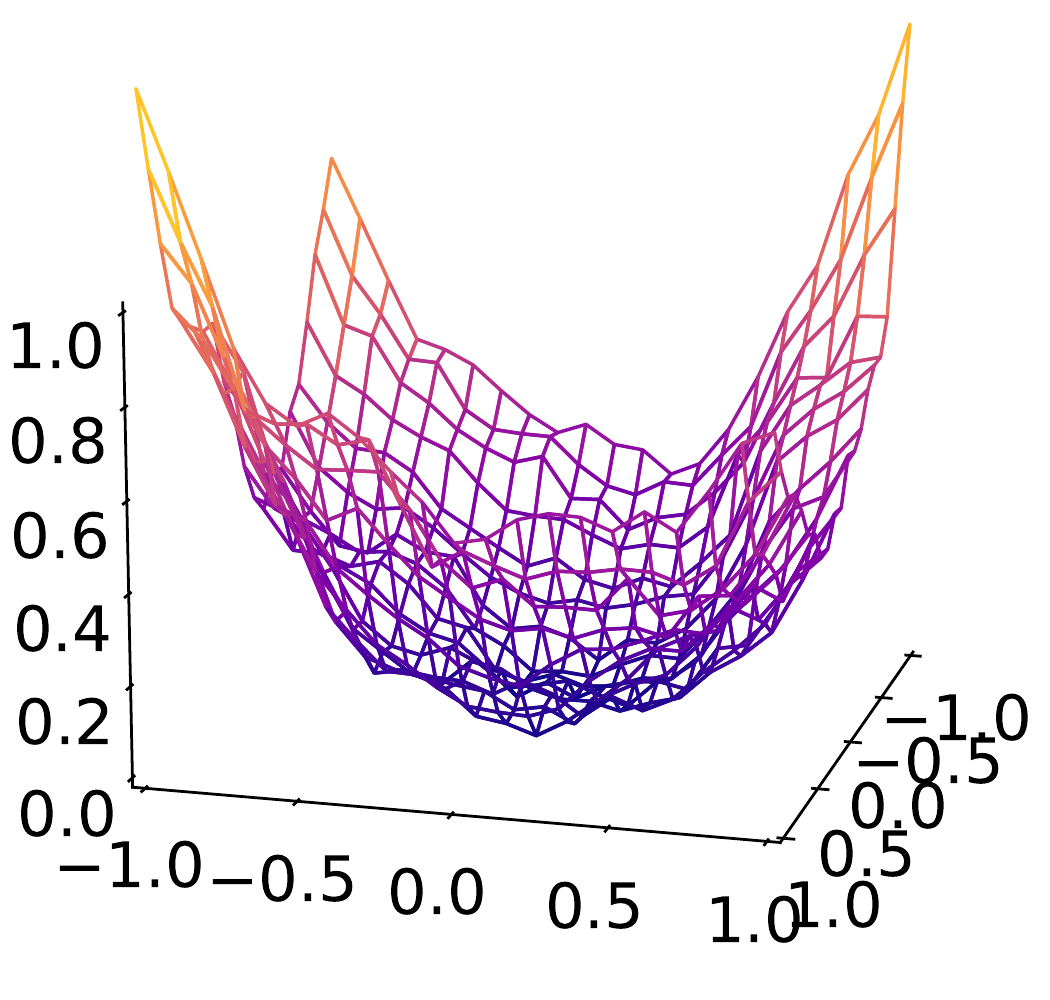}}]{\hspace{0.8cm} wd 0.04$\,\to\,$0.4}
    \makebox[\widthof{\includegraphics[scale=0.25]{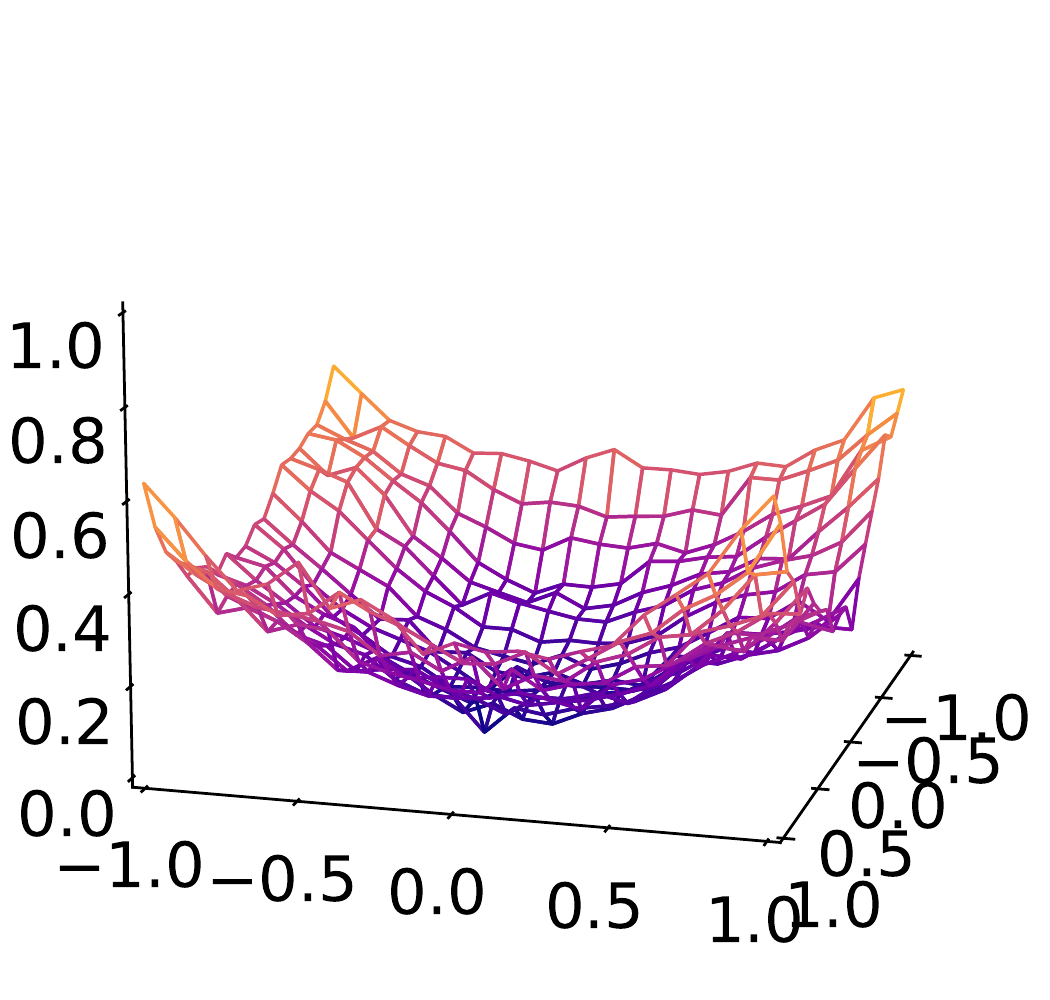}}]{\hspace{0.8cm} wd 0.04$\,\to\,$0.2}
    \par

    \rotatebox{90}{\makebox[\heightof{\includegraphics[scale=0.25]{images/DINO_wd_004_04.pdf}}]{DINO}}
    \hspace{1em}
    \subfigure{\includegraphics[scale=0.25]{images/DINO_wd_004_04.pdf}}
    \subfigure{\includegraphics[scale=0.25]{images/DINO_wd_004_02.pdf}}
    \par

    \rotatebox{90}{\makebox[\heightof{\includegraphics[scale=0.25]{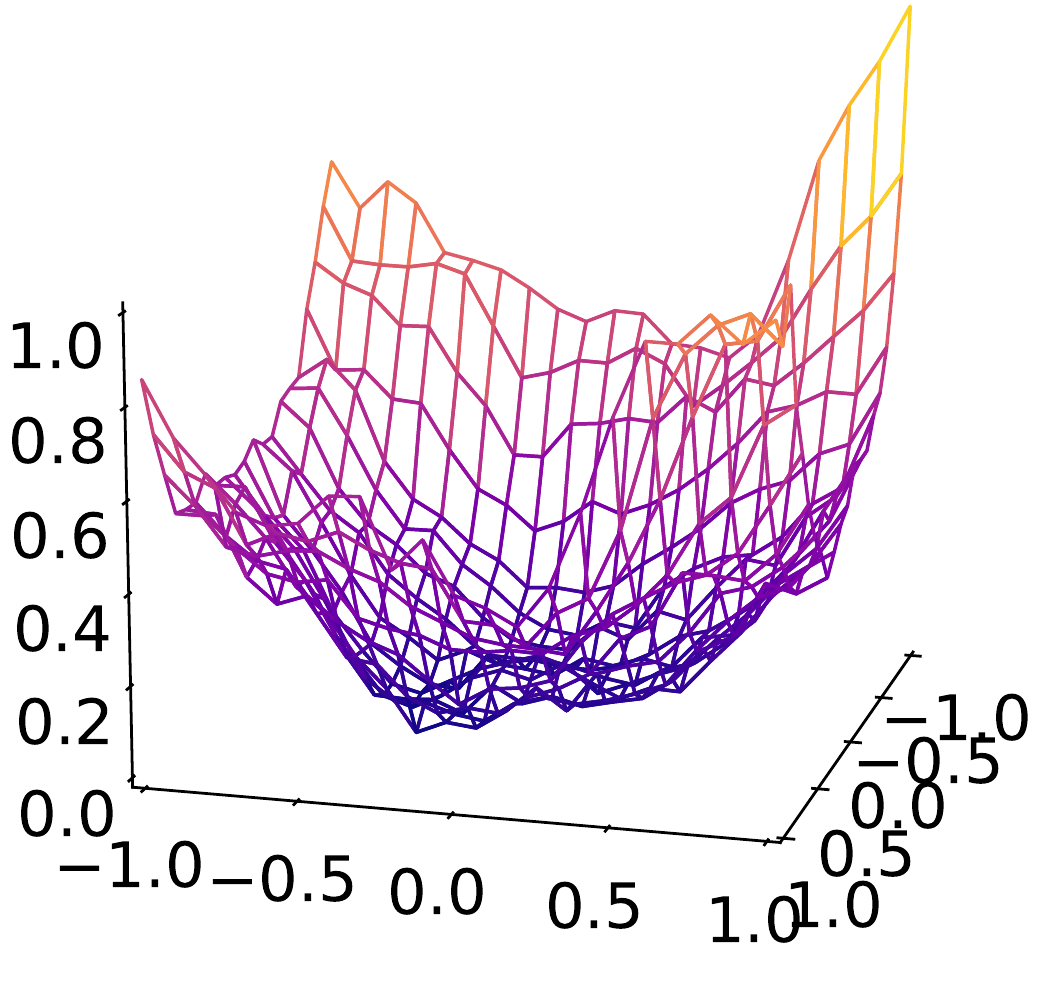}}]{Mugs}}
    \hspace{1em}
    \subfigure{\includegraphics[scale=0.25]{images/Mugs_wd_004_04.pdf}}
    \subfigure{\includegraphics[scale=0.25]{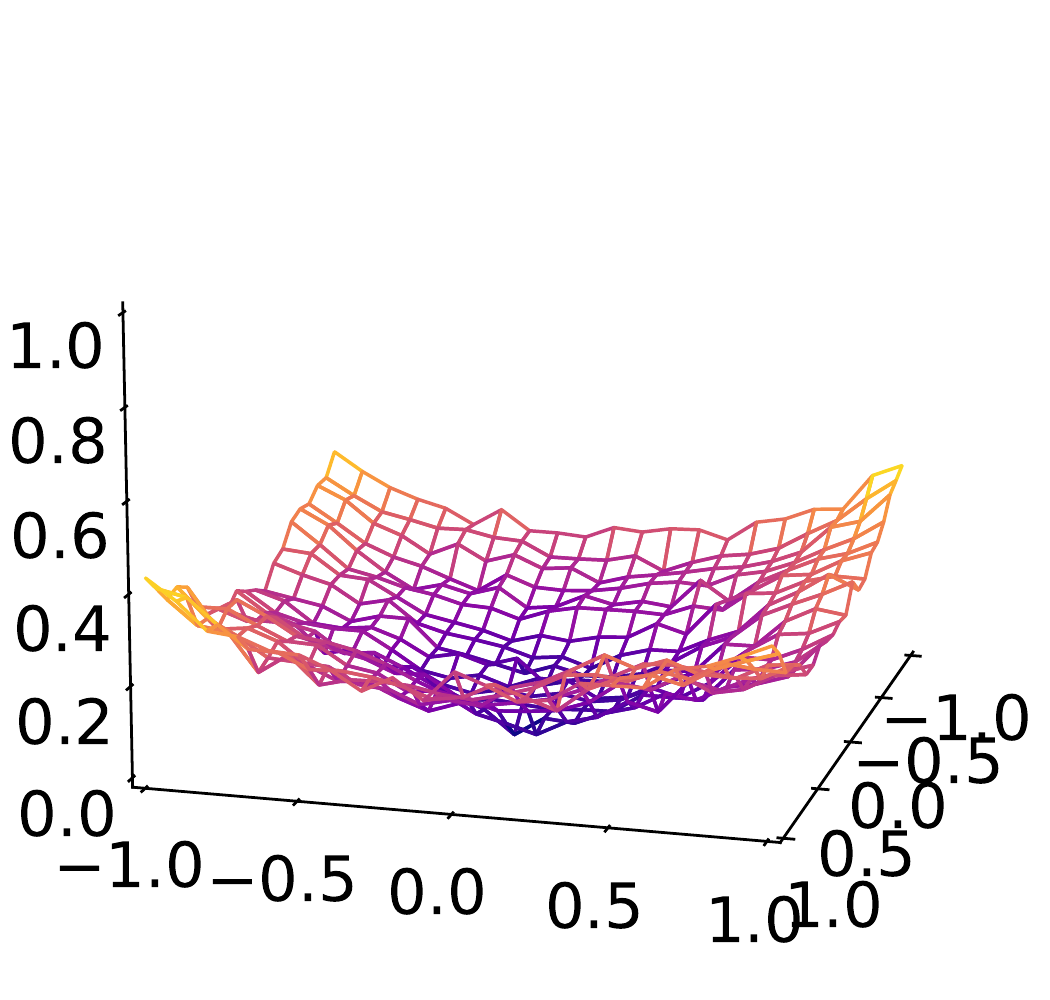}}
\end{center}    
\caption{\textbf{Validation loss landscapes depending on SSL weight decay.} We use the models trained on Stanford Cars in Table~\ref{weight_decay_ablation}. The loss landscape is measured at the end of lr warmup phase of TL. The weight decay of 0.04$\,\to\,$0.2 leads to a flatter loss landscape, indicating the robustness of TL hyperparameter settings.}
\label{fig:loss_landscape}
\end{figure}

\paragraph{Loss landscape.}
\label{loss_landscape}
Through the lens of loss landscape, we aim to investigate how SSL weight decay affects learning behavior in terms of generalization and TL.
We employed the 2D loss landscape visualization method~\cite{VisualizeLoss, HowDoViT} to visualize the loss landscape using the Stanford Cars validation dataset. Specifically, we visualized the checkpoint at the end of the warm-up phase, rather than at the end of training as done in~\cite{HowDoViT}.
Despite having similar validation loss curves, the models trained with the SSL weight decay of 0.04$\,\to\,$0.2 show significantly flatter loss landscapes compared to those with 0.04$\,\to\,$0.4 as can be seen in Figure~\ref{fig:loss_landscape}.
It is well-known that a flatter loss landscape results in improved generalization~\cite{VisualizeLoss, WhenViTResNet, HowDoViT}, and it appears that this is also the case for hyperparameter robustness.
However, it is necessary to decrease SSL weight decay to an appropriate level, as excessively low values of SSL weight decay can make the convergence rate lower as seen in Figure~\ref{fig:loss_curve}. Therefore, we need to set an appropriate value for SSL weight decay to ensure robustness in the TL settings. While it is difficult to observe the robustness of hyperparamters through the loss curve or LP performance, the loss landscape provides good evidence to analyze this.

\begin{figure}[t]
    \centering
        \subfigure[Mugs]{
    \begin{minipage}[t]{0.4\linewidth}
      \centering
      \includegraphics[width=\textwidth]{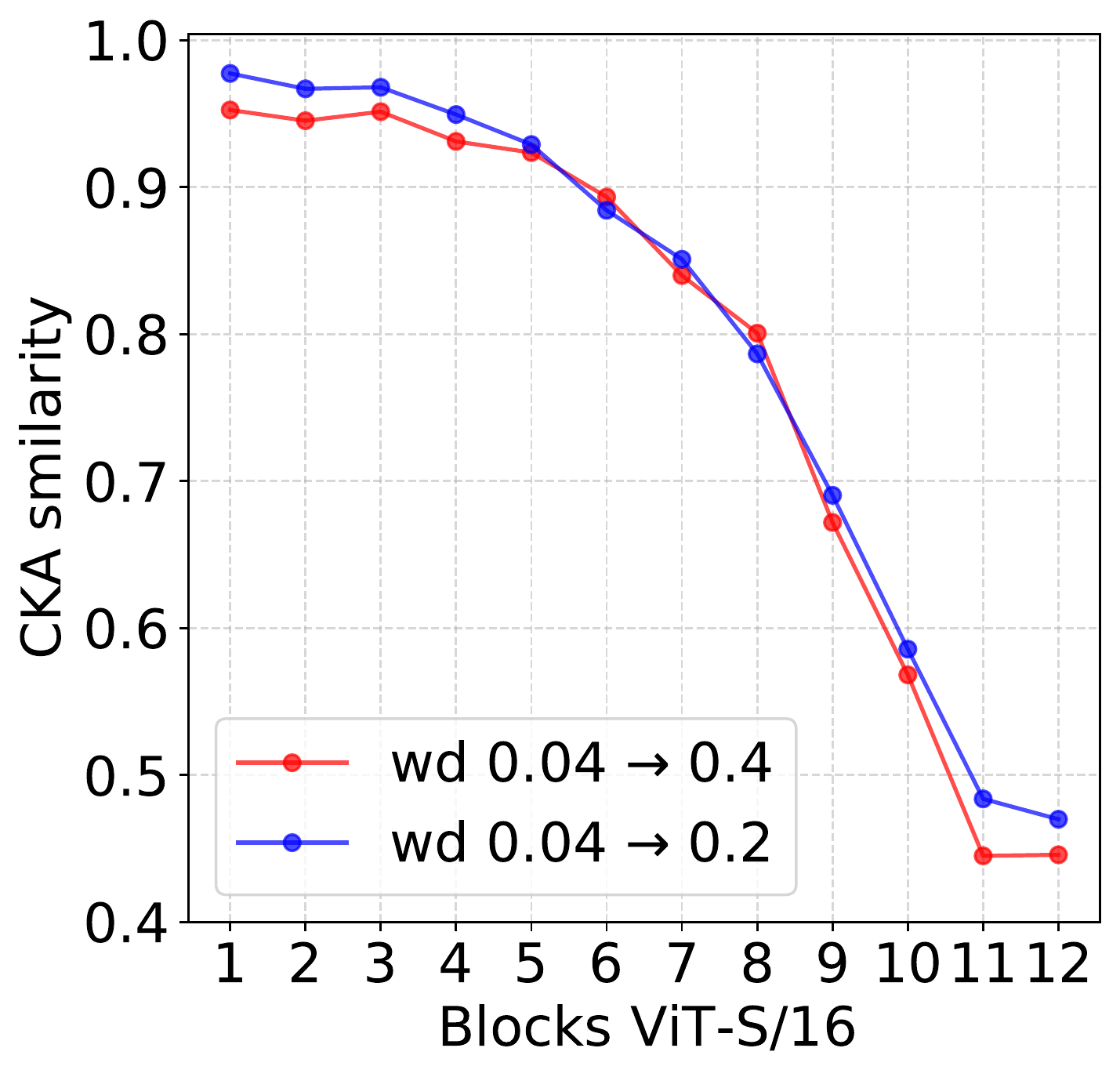}
    \end{minipage}
  }
        \subfigure[DINO]{
    \begin{minipage}[t]{0.4\linewidth}
      \centering
      \includegraphics[width=\textwidth]{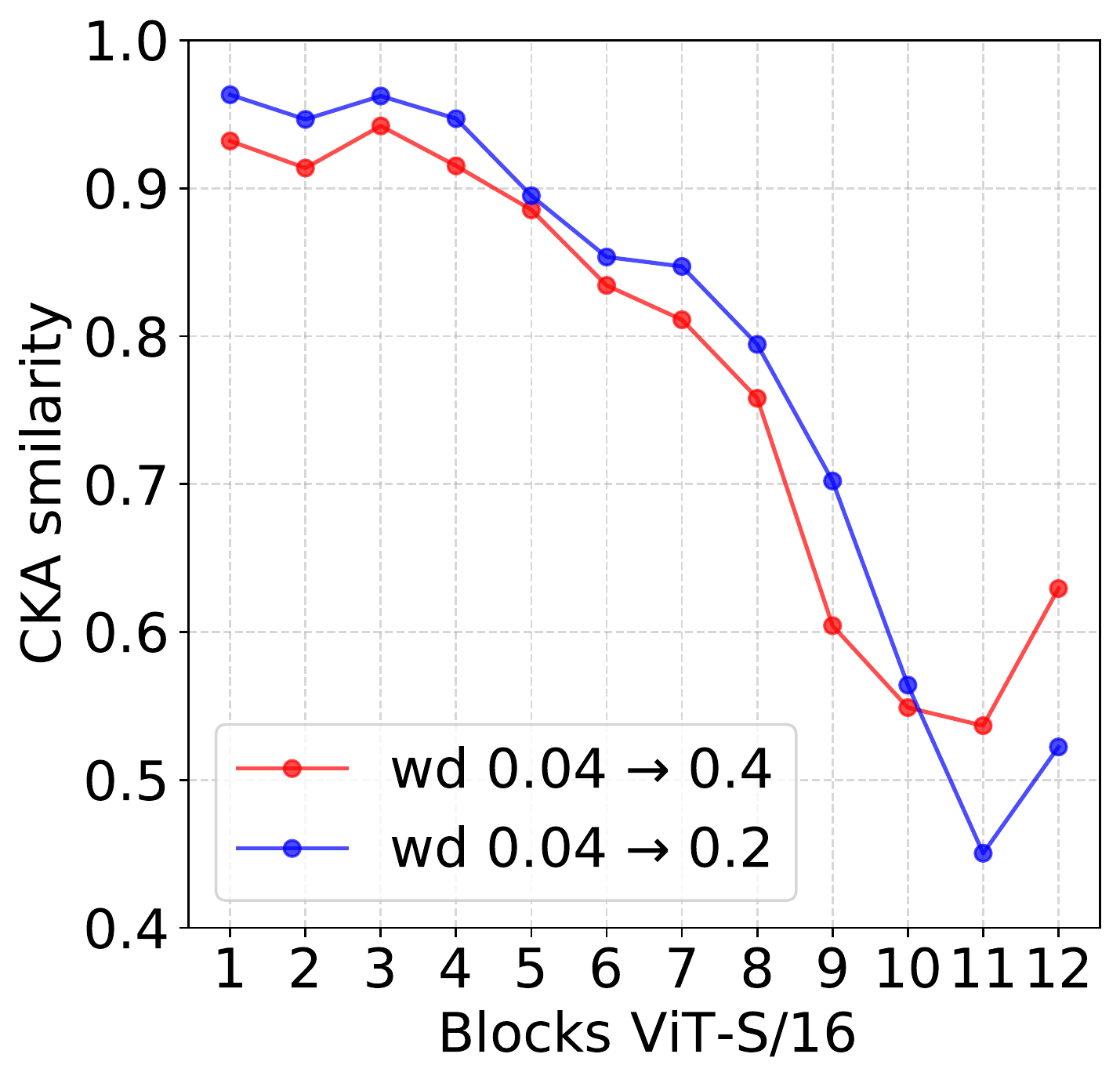}
    \end{minipage}
  }
     \vspace{0pt}
    \caption{
    \textbf{Comparison of CKA values depending on SSL weight decay.} We use the models trained on Stanford Cars in Table~\ref{weight_decay_ablation}. We calculate the CKA similarity of the representations from each block of ViT-S/16 before and after finetuning. The weight decay of 0.04$\,\to\,$0.2 leads to better feature reusability, indicating better transferability.
    }
    \label{fig:cka_wd_abl}
\end{figure}

\paragraph{Feature reusability based on CKA.}
\label{feature_reuse_cka}
In addition, we employed the CKA analysis, a method to intuitively measure the degree of transferability, to analyze the impact of SSL weight decay on feature reusability, which is commonly employed to examine feature similarity in the context of TL. For example, Neyshabur~\textit{et al.}~\cite{WhatIsTransfer} uses CKA similarity to support that if a model has good transferability, features of the model should be reusable.
One of the strengths of TL is that it can improve both training speed and performance by reusing the meaningful representations learned from early--middle layers of a pretrained model~\cite{WhatIsTransfer}. As can be seen in Figure~\ref{fig:cka_wd_abl}, both DINO and Mugs demonstrate an improvement in feature reusability, as measured by CKA, when the SSL weight decay value is set to 0.04$\,\to\,$0.2, particularly in early to middle layers. This observation is consistent with our prior experimental and analytical results, and indicates that an appropriated weight decay value during SSL leads to robust performance across various TL hyperparameters, a flatter loss landscape, and high feature reusability.

\section{Conclusion}
In this work, we address the limitations of commonly used evaluation protocols for SSL representations. Through extensive experiments, we show that existing SSL methods exhibit significant performance variations in both LP and TL depending on the hyperparameters used. In LP, the variations are due to the lack of input normalization in the current evaluation scheme. Interestingly, we find that the cause of performance instability in TL is the weight decay parameter utilized in SSL pretraining, which cannot be detected by discriminability performance metrics like LP or \textit{k}-NN. We believe that our findings shed light on the shortcomings of current evaluation schemes for SSL representations and call for a rethinking of these protocols.

\section*{Acknowledgement}
The author SH was partly supported by the National Research Foundation of Korea (NRF) grant funded by the Korea government (MSIT) (NRF-2021R1C1C1011907).

{\small
\bibliographystyle{ieee_fullname}
\bibliography{egbib}
}

\newpage
\appendix
\onecolumn

\setcounter{figure}{0}
\setcounter{table}{0}
\renewcommand\thefigure{S\arabic{figure}}
\renewcommand\thetable{S\arabic{table}}

\appendix
\label{sec:supplemental}

\section{Datasets}
\label{appendix/datasets}
The ImageNet~\cite{ImageNet} contains 1.2M training images and 50K validation images with 1K classes, widely used as a benchmark dataset for self-supervised learning. CIFAR-100~\cite{CIFAR} is a dataset for multi-class image classification. It consists of 50K training images and 10K test images of 32${\times}$32 resolution with 100 classes. Oxford Flower 102~\cite{Flower} is a dataset for fine-grained image classification. It consists of 2K training images and 6K validation images with 102 flower categories. Each class contains a varying number of images, ranging from 40 to 258. The Stanford Cars~\cite{Car} is a fine-grained dataset including 196 classes of cars. It contains 8K training images and 8K test images. The iNaturalist~\cite{iNat} is a dataset of natural fine-grained categories for image classification. Those categories belong to super-categories. iNaturalist 2018 contains 437K training images and 24K validation images with 8K classes. On the other hand, iNaturalist 2019 consists of 265K training images and 3K validation images with 1K classes, focusing on a smaller set of highly similar categories drawn from iNaturalist.

\section{LP Experimental Details}
\label{appendix/LP_setting}
We examined the LP hyperparameters of popular SSL methods based on the descriptions in each paper and the official code released by the authors, as summarized in Table~\ref{appendix/lp_setting_table}. The ``Concat. ${l}$ last layers'' category indicates the number of outputs from the last layer blocks that are utilized as inputs for the classifier. The ``Patch token'' option allows for the use of the average of patch tokens rather than the class token. The ``BN layer'' option represents whether to use batch normalization for inputs of a classifier. By default, all methods use an input image that is first resized to 256${\times}$256 and then randomly cropped to a size of 224${\times}$224 during training. For inference, images are resized to 256${\times}$256 and center-cropped to a size of 224${\times}$224 for input. All methods do not apply weight decay during LP training.

\begin{table}[H]
\resizebox{\columnwidth}{!}{\begin{tabular}{l|ccccc} \toprule 
 & \begin{tabular}[c]{@{}c@{}} \textbf{DINO / iBOT / Mugs} \\ \textbf{(ViT-S/16)}\end{tabular} & \begin{tabular}[c]{@{}c@{}}\textbf{DINO / iBOT / Mugs}\\ \textbf{(ViT-B/16)}\end{tabular} & \begin{tabular}[c]{@{}c@{}} \textbf{MoCo v3} \\ \textbf{(ViT-S/16, ViT-B/16)}\end{tabular}  & \begin{tabular}[c]{@{}c@{}} \textbf{MAE} \\ \textbf{(ViT-B/16)}\end{tabular} & \begin{tabular}[c]{@{}c@{}} \textbf{MSN} \\ \textbf{(ViT-S/16, ViT-B/16)}\end{tabular} \\ \midrule
\textbf{Epochs} & 100 & 100 & 90 & 90 & 100 \\
\textbf{Batch size} & 1024 & 1024 & 4096 & 16384 & 16384 \\
\textbf{Optimizer} & SGD & SGD & SGD & LARS & SGD \\
\textbf{Learning rate (LR)} & 0.001 / 0.001 / 0.04 &  0.001 / 0.001 / 0.008 & 3 & 0.1 & 6.4 \\
\textbf{LR warm-up epochs} & \xmark & \xmark & \xmark & 10 epoch & \xmark \\
\textbf{LR decay} & CosineAnnealingLR & CosineAnnealingLR & CosineAnnealingLR & CosineAnnealingLR & CosineAnnealingLR \\
\textbf{Weight decay} & \xmark & \xmark & \xmark & \xmark & \xmark \\
\textbf{Concat. ${l}$ last layers} & 4 & 1 & 1 & 1 & 1 \\
\textbf{Patch token} & \xmark & \cmark & \xmark & \xmark & \xmark \\
\textbf{BN layer} & \xmark & \xmark & \xmark & \cmark & \cmark \\ \bottomrule
\end{tabular}}
\vspace{1pt}
\caption{\textbf{LP settings for the SSL methods}. We can observe that the LP hyperparameter settings of each method are quite different. Note that all SSL methods conduct extensive hyperparameter searches to obtain these optimal configurations.
}
\label{appendix/lp_setting_table}
\end{table}

\begin{table*}[h]
    \centering
    \scalebox{0.8}{
    \begin{tabular}{l|cccccc} 
        \toprule
        & \multicolumn{6}{c}{\textbf{\textit{k}-NN}} \\
        \cline{2-7}
        \textbf{Method} & \textbf{Paper} & \textbf{\textit{k}=5} & \textbf{\textit{k}=10} & \textbf{\textit{k}=20} & \textbf{\textit{k}=50} & \textbf{\textit{k}=100} \\ 
        \midrule
       \textbf{SL (DeiT)} & 79.8 & 78.6 & 79.2 & \textbf{79.3} & 79.2 & 79.1 \\
        \textbf{DINO} & 74.5 & 73.8 & \textbf{74.4} & \textbf{74.4} & 73.7 & 72.9 \\
        \textbf{MoCo v3} & - & 66.9 & 68.0 & \textbf{68.2} & 67.3 & 66.4 \\
        \textbf{Mugs} & 75.6 & 74.8 & \textbf{75.5} & 75.4 & 74.8 & 73.9 \\
        \textbf{iBOT} & 75.2 & 74.4 & \textbf{75.2} & 74.9 & 74.3 & 73.6 \\
        \textbf{MSN} & - & 74.3 & 74.7 & \textbf{74.8} & 74.0 & 73.1 \\
        \bottomrule
    \end{tabular}}
    \vspace{5pt}
    \caption{\textbf{ImageNet \textit{k}-NN classification accuracy}. Best accuracy for each method is shown in bold.}
\label{table:imagenet_knn}
\end{table*}

\section{\textit{k}-NN Results on ImageNet}
\label{appendix/kNN_result}
\textit{k}-NN is a frequently used classifier for evaluating discriminability. Specifically, the performance of \textit{k}-NN classifiers is evaluated based on the features obtained from SSL backbones that are frozen. 
\textit{k}-NN has only a single hyperparameter, i.e., the value of \textit{k}.
As shown in Table~\ref{table:imagenet_knn}, although there are performance differences depending on the \textit{k} value, these differences are very small. Based on the official ViT-S/16 checkpoints of various SSL methods, we observed that the best performance is achieved when the value of \textit{k} is set to 10 or 20. Since both values of \textit{k} yield similar results, we chose 20 as the default \textit{k} value for all experiments.

\section{TL Experimental Details}
\label{appendix/TL_setting}
We report the hyperparameters setting of SL/SSL methods for TL in Table~\ref{appendix/ft_setting_table}. 
We used the hyperparameter settings described in the literature as a basis. For values that are not explicitly mentioned in the papers, we referred to the official codes. For unknown values, e.g., the random erasing probability of Mugs, we marked them as `?'. Mugs did not specify exact learning rate values except for iNat18 and iNat19, but mentioned that those values were chosen by searching within the set \{7.50e-6, 1.50e-5, 3.00e-5, 7.50e-5, 1.50e-4\}.
The TL hyperparameter setting for Stanford Cars is not provided in the MoCo v3 paper. Considering that the TL settings for Stanford Cars and Flower-102 are the same in other SSL methods such as DINO and iBOT, the Flower-102 setting was used for the Stanford Cars experiments of MoCo v3. Note that many methods (e.g. DINO, Mugs, iBOT, and MSN) follow the hyperparameter settings of DeiT. 
That is, most methods are employing strong regularization techniques (e.g. label smoothing~\cite{LabelSmoothing}, repeated augmentation~\cite{RepeatedAugmentation}, Mixup~\cite{Mixup}, CutMix~\cite{Cutmix}, and random erasing~\cite{ErasingProb}), which boost TL performance regardless of the quality of representations. Therefore, in the ``Short'' TL setting, we exclude all of these regularizations and set a short training epoch to diminish the performance-boosting effect, thereby enabling us to assess the transferability of these SSL backbones.

\newpage
\begin{table}[hb]
\begin{center}
\begin{small}
\resizebox*{!}{.24\textheight}{

\begin{tabular}{l|cc|cccccc}
\hline
 & \multicolumn{2}{c|}{\textbf{DeiT}} & \multicolumn{6}{c}{\textbf{DINO}} \\ \hline
\textbf{Downstream dataset} & \textbf{CIFAR-10/100} & \textbf{Cars} & \textbf{CIFAR-10} & \textbf{CIFAR-100} & \textbf{iNat18} & \textbf{iNat19} & \textbf{Flowers} & \textbf{Cars} \\ \hline
\textbf{Epochs} & 1000 & 1000 & 1000 & 1000 & 300 & 300 & 1000 & 1000  \\
\textbf{Batch size} & 768 & 768 & 768 & 768 & 1024 & 1024 & 768 & 768  \\
\textbf{Optimizer} & SGD & SGD & SGD & AdamW & AdamW & AdamW & AdamW & AdamW \\
\textbf{Learning rate (LR)} & 1.00e-02 & 1.00e-02 & 5.00e-06 & 5.00e-06 & 5.00e-05 & 5.00e-05 & 5.00e-06 & 5.00e-06 \\
\textbf{LR decay} & cosine & cosine & cosine & cosine & cosine & cosine & cosine & cosine \\
\textbf{LR warm-up epochs} & 5 & 5 & 5 & 5 & 5 & 5 & 5 & 5 \\
\textbf{LR warm-up} & 1.00e-06 & 1.00e-06 & 1.00e-06 & 1.00e-06 & 1.00e-06 & 1.00e-06 & 1.00e-06 & 1.00e-06 \\ 
\textbf{Weight decay} & 1.00e-04 & 1.00e-04 & 0.05 & 0.05 & 0.05 & 0.05 & 0.05 & 0.05 \\ \hline
\textbf{Label smoothing} & 0.1 & 0.1 & 0.1 & 0.1 & 0.1 & 0.1 & 0.1 & 0.1 \\
\textbf{Drop path rate} & 0 & 0 & 0.1 & 0.1 & 0.1 & 0.1 & 0.1 & 0.1 \\
\textbf{Repeated Aug.} & \cmark & \cmark & \cmark & \cmark & \cmark & \cmark & \cmark & \cmark \\
\textbf{Rand Aug.} & 9 / 0.5 & 9 / 0.5 & 9 / 0.5 & 9 / 0.5 & 9 / 0.5 & 9 / 0.5 & 9 / 0.5 & 9 / 0.5 \\
\textbf{Mixup prob.} & 0.8 & 0.8 & 0.8 & 0.8 & 0.8 & 0.8 & 0.8 & 0.8 \\
\textbf{CutMix prob.} & 1 & 1 & 1 & 1 & 1 & 1 & 1 & 1 \\
\textbf{Erasing Prob.} & \xmark & \xmark & \xmark & \xmark & 0.25 & 0.25 & \xmark & \xmark \\ \hline

\end{tabular}}
\end{small}
\end{center}
\label{appendix/ft_setting_table_1}

\vspace{-15pt}
\begin{center}
\begin{small}
\resizebox*{!}{.24\textheight}{
\begin{tabular}{l|cccc|ccccc}
\hline
 & \multicolumn{4}{c|}{\textbf{MoCoV3}} & \multicolumn{5}{c}{\textbf{Mugs}} \\ \hline
\textbf{Downstream dataset} & \textbf{ImageNet} & \textbf{CIFAR-10} & \textbf{CIFAR-100} & \textbf{Flowers} & \textbf{CIFAR-10/100} & \textbf{iNat18} & \textbf{iNat19} & \textbf{Flowers} & \textbf{Cars} \\ \hline
\textbf{Epochs} & 150 & 100 & 100 & 100 & 1000 & 360 & 360 & 1000 & 1000 \\
\textbf{Batch size} & 1024 & 1024 & 1024 & 1024 & 768 & 768 & 768 & 768 & 768 \\
\textbf{Optimizer} & AdamW & AdamW & AdamW & AdamW & AdamW & AdamW & AdamW & AdamW & AdamW \\
\textbf{Learning rate (LR)} & 5.00e-04 & 3.00e-04 & 3.00e-04 & 3.00e-04 & ? & 3.00e-05 & 7.50e-05 & ? & ? \\
\textbf{LR decay} & cosine & cosine & cosine & cosine & cosine & cosine & cosine & cosine & cosine \\
\textbf{LR warm-up epochs} & 3 & 3 & 3 & 3 & 5 & 5 & 5 & 5 & 5 \\
\textbf{LR warm-up} & 1.00e-06 & 1.00e-06 & 1.00e-06 & 1.00e-06 & 1.00e-06 & 1.00e-06 & 1.00e-06 & 1.00e-06 & 1.00e-06 \\ 
\textbf{Weight decay} & 0.05 & 0.1 & 0.1 & 0.1 & 0.05 & 0.05 & 0.05 & 0.05 & 0.05 \\ \hline
\textbf{Label smoothing} & 0.1 & 0.1 & 0.1 & 0.1 & 0.1 & 0.1 & 0.1 & 0.1 & 0.1 \\
\textbf{Drop path rate} & 0.1 & 0.1 & 0.1 & 0.1 & 0.1 & 0.1 & 0.1 & 0.1 & 0.1 \\
\textbf{Repeated Aug.} & \cmark & \cmark & \cmark & \cmark & \cmark & \cmark & \cmark & \cmark & \cmark \\
\textbf{Rand Aug.} & 9 / 0.5 & 9 / 0.5 & 9 / 0.5 & 9 / 0.5 & 9 / 0.5 & 9 / 0.5 & 9 / 0.5 & 9 / 0.5 & 9 / 0.5 \\
\textbf{Mixup prob.} & 0.8 & 0.8 & 0.5 & 0 & 0.8 & 0.8 & 0.8 & 0.8 & 0.8 \\
\textbf{CutMix prob.} & 1 & 1 & 1 & 0 & 1 & 1 & 1 & 1 & 1 \\
\textbf{Erasing Prob.} & 0.25 & \xmark & \xmark & 0.25 & ? & ? & ? & ? & ? \\ \hline

\end{tabular}}
\end{small}
\end{center}
\label{appendix/ft_setting_table_2}
\vspace{-15pt}
\begin{center}
\begin{small}
\resizebox*{!}{.24\textheight}{
\begin{tabular}{l|ccccc|cccc|c}
\hline
 & \multicolumn{5}{c|}{\textbf{iBOT}} & \multicolumn{4}{c|}{\textbf{MSN}} & \textbf{Short} \\ \hline
\textbf{Downstream dataset} & \textbf{CIFAR-10/100} & \textbf{iNat18} & \textbf{iNat19} & \textbf{Flowers} & \textbf{Cars} & \textbf{CIFAR-10} & \textbf{CIFAR-100} & \textbf{iNat18} & \textbf{iNat19} & \textbf{All} \\ \hline
\textbf{Epochs} & 1000 & 360 & 360 & 1000 & 1000 & 1000 & 1000 & 300 & 300 & 50 \\
\textbf{Batch size} & 768 & 768 & 768 & 768 & 768 & 768 & 768 & 1024 & 1024 & 64 \\
\textbf{Optimizer} & AdamW & AdamW & AdamW & AdamW & AdamW & SGD & AdamW & AdamW & AdamW & Adam \\
\textbf{Learning rate (LR)} & 7.50e-06 & 5.00e-05 & 2.50e-05 & 7.50e-06 & 7.50e-06 & 7.50e-05 & 7.50e-05 & 1.00e-04 & 1.00e-04 & 1.00e-04 \\
\textbf{LR decay} & cosine & cosine & cosine & cosine & cosine & cosine & cosine & cosine & cosine & step \\
\textbf{LR warm-up epochs} & 5 & 5 & 5 & 5 & 5 & 5 & 5 & 5 & 5 & \xmark \\
\textbf{LR warm-up} & 1.00e-06 & 1.00e-06 & 1.00e-06 & 1.00e-06 & 1.00e-06 & 1.00e-06 & 1.00e-06 & 1.00e-06 & 1.00e-06 & \xmark \\
\textbf{Weight decay} & 0.05 & 0.05 & 0.05 & 0.05 & 0.05 & 0.05 & 0.05 & 0.05 & 0.05 & 1.00e-06 \\ \hline
\textbf{Label smoothing} & 0.1 & 0.1 & 0.1 & 0.1 & 0.1 & 0.1 & 0.1 & 0.1 & 0.1 & \xmark \\
\textbf{Drop path rate} & 0.1 & 0.1 & 0.1 & 0.1 & 0.1 & 0.1 & 0.1 & 0.1 & 0.1 & 0.1 \\
\textbf{Repeated Aug.} & \cmark & \cmark & \cmark & \cmark & \cmark & \cmark & \cmark & \cmark & \cmark & \xmark \\
\textbf{Rand Aug.} & 9 / 0.5 & 9 / 0.5 & 9 / 0.5 & 9 / 0.5 & 9 / 0.5 & 9 / 0.5 & 9 / 0.5 & 9 / 0.5 & 9 / 0.5 & \xmark \\
\textbf{Mixup prob.} & 0.8 & 0.8 & 0.8 & 0.8 & 0.8 & 0.8 & 0.8 & 0.8 & 0.8 & \xmark \\
\textbf{CutMix prob.} & 1 & 1 & 1 & 1 & 1 & 1 & 1 & 1 & 1 & \xmark \\
\textbf{Erasing Prob.} & 0.25 & 0.1 & 0.1 & 0.25 & 0.25 & \xmark & \xmark & 0.25 & 0.25 & \xmark \\ \hline
\end{tabular}}
\end{small}
\end{center}
\caption{\textbf{TL settings for SL/SSL methods on ViT-S/16.} The TL hyperparameter settings of each method and dataset are considerably different.
}
\label{appendix/ft_setting_table}
\end{table}

\section{Stability Analysis on Mugs~\cite{Mugs}}
\label{mugs_abl}

Here, we present the results of an ablation study on Mugs~\cite{Mugs}. We initially checked if a particular Mugs module contributes to stable TL performance over DINO. Subsequently, we investigated the components of SSL training settings that affect the transferability of SSL representations. Lastly, we conducted experiments to determine the influence of the hyperparameters utilized in the SSL stage on transferability.

\newpage
\subsection{Module analysis} 
The main contribution of Mugs is its ability to learn multi-granular features based on instance, local group, and group modules. Consequently, it is presumed that the newly introduced module is the most significant factor affecting stability. In the literature, the authors conducted a module removal ablation study concerning LP. In this section, we further examine the impact of these modules on TL performance. As shown in Table~\ref{table:mugs_module}, while the LP and TL performances exhibit slight variations depending on hyperparameter settings, there is little difference in stability, except when solely using the local group module. This is due to the issue of utilizing less-trained features as neighbors when using the local group module alone, leading to a reduction in performance.
Note that using the instance module alone can be viewed as an improved version of MoCo v2 based on ViT, but it differs from MoCo v3 in terms of memory for negative samples. Similarly, using the group module alone has exactly the same architecture as DINO. Based on the ablation results, it can be observed that Mugs shows stable TL performance even when trained only with the group module, which has the same model architecture as DINO. Therefore, we can infer that other factors contribute to the stable TL performance of Mugs.

\begin{table}[H]
    \centering
    \scalebox{0.9}{
    \begin{tabular}{ccc|ccc|ccc} 
        \toprule
        \multicolumn{3}{c|}{\textbf{Mugs module}} & \multicolumn{3}{c|}{\textbf{LP setting}} & \multicolumn{3}{c}{\textbf{TL setting}} \\
        \cline{1-9}
        \textbf{Instance} & \textbf{Local group} & \textbf{Group} & \textbf{DINO} & \textbf{MoCo v3} & \textbf{MAE} & \textbf{DINO} & \textbf{MoCo v3} & \textbf{Short} \\ 
        \midrule
        \cmark  & \cmark  & \cmark  & 74.97 & 74.39 & 74.44 & 91.21 & 88.78 & 74.04 \\
        \xmarkg & \xmarkg & \cmark  & 74.49 & 73.79 & 73.84 & 90.78 & 88.37 & 72.31 \\
        \xmarkg & \cmark  & \xmarkg & \textit{\underline{31.33}} & \textit{\underline{59.46}} & \textit{\underline{66.36}} & \textit{\underline{89.96}} & \textit{\underline{79.32}} & \textit{\underline{47.56}} \\
        \xmarkg & \cmark  & \cmark  & 75.04 & 74.35 & 74.52 & 91.16 & 88.75 & 75.76 \\
        \cmark  & \xmarkg & \xmarkg & 74.50 & 73.78 & 73.84 & 90.70 & 88.45 & 74.55 \\
        \cmark  & \xmarkg & \cmark  & 75.43 & 74.61 & 74.67 & 90.92 & 88.79 & 74.92 \\
        \cmark  & \cmark  & \xmarkg & 74.40 & 73.84 & 73.85 & 91.11 & 88.46 & 77.05 \\
        \bottomrule
    \end{tabular}}
    \vspace{5pt}
    \caption{\textbf{Effect of Mugs modules.} Each model is trained for 100 epochs with ViT-S/16. We report LP results on ImageNet and TL results on Stanford Cars. Except for the local group module only, there is no difference in terms of stability. Unstable results are underlined.}
\label{table:mugs_module}
\end{table}

\subsection{Ablation study on training settings: Mugs vs. DINO}
In order to determine where the stability difference between `DINO' and `Mugs with the group module only' originates, we conducted additional analyses related to the training settings. Through a comparison of DINO and Mugs' official codes, we discovered several differences in detail, including augmentation, weight normalization, and SSL hyperparameters. When comparing the augmentation schemes for DINO and Mugs, they have nearly identical configurations, but Mugs incorporates an extra strong augmentation technique known as RandAugment~\cite{RandAugment}. In terms of weight normalization, DINO employs weight normalization for the teacher network's last layer, whereas Mugs does not. Moreover, Mugs and DINO have different SSL hyperparameters.

We sampled some combinations of settings for the ablation study because it would take too much time and resources to consider all combinations. Based on experimental results, as shown in Table~\ref{table:ssl_setting_change_results}, we noticed that DINO's stability (i.e., transferability) considerably increased when trained under Mugs' SSL hyperparameters while retaining all other settings identical to the original DINO.

\begin{table}[H]
    \centering
    \scalebox{0.9}{
    \begin{tabular}{c|ccc|ccc|ccc}
    \toprule
    \multirow{2}{*}{\textbf{Base}} & \multicolumn{3}{c|}{\textbf{SSL setting}} & \multicolumn{3}{c|}{\textbf{LP setting}} & \multicolumn{3}{c}{\textbf{TL setting}} \\ 
    \cline{2-10}
    & \textbf{Augmentation} & \textbf{Teacher WN} & \textbf{Hyperparams.} & \textbf{DINO} & \textbf{MoCo} & \textbf{MAE} & \textbf{DINO} & \textbf{MoCo V3} & \textbf{Short} \\
    \midrule
    \multirow{3}{*}{Mugs} & Mugs & Mugs & Mugs & 74.49 & 73.79 & 73.84 & 90.78 & 88.37 & 72.31 \\
    & DINO & DINO & DINO & 74.58 & 73.21 & 73.49 & \textit{\underline{91.29}} & \textit{\underline{84.47}} & \textit{\underline{43.29}} \\
    & DINO & DINO & Mugs & 74.03 & 73.43 & 73.56 & 90.44 & 88.87 & 72.82 \\
    \midrule
    \multirow{2}{*}{DINO} & DINO & DINO & DINO & 74.12 & 72.62 & 72.89 & \textit{\underline{91.26}} & \textit{\underline{24.01}} & \textit{\underline{19.93}} \\
    & DINO & DINO & Mugs & 74.30 & 73.78 & 73.94 & 90.49 & 87.23 & 67.15 \\ 
    \bottomrule
    \end{tabular}}
    \vspace{5pt}
    \caption{\textbf{Ablation study on Mugs and DINO training settings.} Each model is trained for 100 epochs with ViT-S/16. We report LP results on ImageNet and TL results on Stanford Cars. While the SSL hyperparameters of Mugs show stable results in both Mugs and DINO, those of DINO produce unstable results in both Mugs and DINO. Conversely, augmentations and teacher weight normalization do not appear to have a significant impact. Unstable results are underlined.}
    \label{table:ssl_setting_change_results}
\end{table}

\subsection{Ablation study on SSL hyperparameters: Mugs vs. DINO} 
We aim to analyze which elements of SSL hyperparameters have the greatest influence on transferability. Upon comparing DINO and Mugs, we discovered several distinct SSL hyperparameters, such as patch embedding learning rate, teacher softmax temperature, gradient clipping, learning rate, and weight decay.
Note that hyperparameter settings differ depending on the network architecture. In this section, we investigate the SSL hyperparameters for ViT-S/16. For learning rate, both DINO and Mugs employ cosine scheduling and linear warmup strategies. The learning rate value of DINO is 5e-4$\,\to\,$1e-5, while that of Mugs is 8e-4$\,\to\,$1e-6. Similarly, cosine scheduling is used for weight decay. The weight decay value of DINO is 0.04$\,\to\,$0.4, and that of Mugs is 0.04$\,\to\,$0.2. Only Mugs incorporates a 20\% reduction in patch embedding learning rate. DINO does not use gradient clipping, but Mugs use it with a threshold of 3.0. Lastly, DINO schedules the softmax temperature applied to the teacher's outputs from 0.04$\,\to\,$0.07, while Mugs employs a fixed value of 0.04.

As shown in Table~\ref{table:appendix/dino_mugs_abl_results}, it can be observed that the weight decay parameter in the SSL stage has a significant impact on the transferability of SSL representations, i.e., adjusting the weight decay value from 0.04$\,\to\,$0.4 to 0.04$\,\to\,$0.2 considerably stabilizes DINO's TL performance. 
Note that similar to the previous experiments, we sampled some combinations instead of considering all possible cases. It is worth noting that it is difficult to identify such an influence on TL performance based on the perspective of discriminability evaluated by LP.

\begin{table}[H]
\begin{center}
\begin{small}
\resizebox{0.65\columnwidth}{!}{\begin{tabular}{c|ccccc|cc|ccc} \toprule
\textbf{Base setting} & \multicolumn{5}{c|}{\textbf{SSL HP setting}} & \multicolumn{2}{c|}{\textbf{LP setting}} & \multicolumn{3}{c}{\textbf{TL setting}} \\ \midrule
 & \rotatebox{90}{\textbf{Patch embedding LR}} & \rotatebox{90}{\textbf{Gradient clipping}} & \rotatebox{90}{\textbf{Learning rate}} & \rotatebox{90}{\textbf{Weight decay}} & \rotatebox{90}{\textbf{Teacher temperature}} & \rotatebox{90}{\textbf{DINO}} & \rotatebox{90}{\textbf{MoCo v3}} & \rotatebox{90}{\textbf{DINO}} & \rotatebox{90}{\textbf{MoCo v3}} & \rotatebox{90}{\textbf{Short}} \\ \midrule
\multirow{6}{*}{Mugs} & M & M & M & M & M & 74.03 & 73.43 & 90.44 & 88.87 & 72.82 \\
 & D & D & D & D & D & 74.58 & 73.21 & \textit{\underline{91.29}} & \textit{\underline{84.47}} & \textit{\underline{43.29}} \\
 & D & M & M & M & M & 73.52 & 72.90 & 90.56 & 89.72 & 80.45 \\
 & M & D & M & M & M & 73.84 & 73.16 & 90.93 & 88.54 & 75.95 \\
 & M & M & D & M & M & 75.16 & 74.22 & 91.75 & 88.50 & 70.87 \\
 & M & M & M & D & M & 74.94 & 73.74 & \textit{\underline{91.48}} & \textit{\underline{80.80}} & \textit{\underline{44.99}} \\ \midrule
\multirow{10}{*}{DINO} & D & D & D & D & D & 74.12 & 72.62 & \textit{\underline{91.26}} & \textit{\underline{24.01}} & \textit{\underline{19.93}} \\
 & M & M & M & M & M & 74.30 & 73.78 & 90.49 & 87.23 & 67.15 \\
 & M & M & D & D & M & 74.78 & 73.65 & \textit{\underline{91.58}} & \textit{\underline{9.45}} & \textit{\underline{13.34}} \\
 & M & M & D & D & D & 74.64 & 73.36 & \textit{\underline{91.47}} & \textit{\underline{25.67}} & \textit{\underline{15.90}} \\
 & D & D & M & M & M & 73.57 & 73.03 & 90.32 & 87.84 & 71.42 \\
 & D & D & M & M & D & 73.56 & 73.10 & 90.06 & 88.52 & 77.19 \\
 & M & M & M & M & D & 74.23 & 73.69 & 90.60 & 88.38 & 63.28 \\
 & D & D & D & D & M & 74.12 & 72.79 & \textit{\underline{91.13}} & \textit{\underline{24.47}} & \textit{\underline{36.37}} \\
 & D & D & M & D & D & 73.05 & 71.95 & \textit{\underline{90.98}} & \textit{\underline{11.67}} & \textit{\underline{34.45}} \\
 & D & D & D & M & D & 73.65 & 73.07 & 90.80 & 87.24 & 73.39 \\ \bottomrule
\end{tabular}}
\end{small}
\end{center}
\caption{\textbf{Impact of hyperparameters in SSL upstream pretraining.} D and M represent the settings of DINO and Mugs, respectively. It can be observed that SSL weight decay determines the robustness of TL performance. Unstable results are underlined.}
\label{table:appendix/dino_mugs_abl_results}
\end{table}

\end{document}